\documentclass[journal]{IEEEtran}
\usepackage{amsmath}
\usepackage{cases}
\usepackage{amssymb}
\usepackage{graphicx}
\usepackage{subfigure}
\usepackage{algorithmic}
\usepackage{algorithm}
\usepackage{float}
\usepackage{booktabs}
\usepackage{textcomp}
\usepackage{array}
\ifCLASSINFOpdf

\else

\fi

\begin{document}

\title{Similarity and Dissimilarity Guided Co-association Matrix Construction for Ensemble Clustering}

\author{Xu~Zhang, Yuheng~Jia,~\IEEEmembership{Member,~IEEE,}
Mofei~Song,~\IEEEmembership{Member,~IEEE,}, Ran~Wang~\IEEEmembership{Senior Member}
        
\thanks{X. Zhang. Y. Jia and M. Song are with School of Computer Science and Engineering, South University, Nanjing 210093, China (email: xuz@seu.edu.cn, yhjia@seu.edu.cn, songmf@seu.edu.cn). 
R. Wang is with the School of Mathematical Sciences, Shenzhen University, Shenzhen 518060, China, with the Guangdong Provincial Key Laboratory of Intelligent Information Processing, Shenzhen University, Shenzhen 518060, China, and also with the Shenzhen Key Laboratory of Advanced Machine Learning and Applications, Shenzhen University, Shenzhen 518060, China. (e-mail: wangran@szu.edu.cn).}
}

\maketitle

\begin{abstract}
Ensemble clustering aggregates multiple weak clusterings to achieve a more accurate and robust consensus result. The Co-Association matrix (CA matrix) based method is the mainstream ensemble clustering approach that constructs the similarity relationships between sample pairs according the weak clustering partitions to generate the final clustering result. However, the existing methods neglect that the quality of cluster is related to its size, i.e., a cluster with smaller size tends to higher accuracy. Moreover, they also do not consider the valuable dissimilarity information in the base clusterings which can reflect the varying importance of sample pairs that are completely disconnected. To this end, we propose the Similarity and Dissimilarity Guided Co-association matrix (SDGCA) to achieve ensemble clustering. First, we introduce normalized ensemble entropy to estimate the quality of each cluster, and construct a similarity matrix based on this estimation. Then, we employ the random walk to explore high-order proximity of base clusterings to construct a dissimilarity matrix. Finally, the adversarial relationship between the similarity matrix and the dissimilarity matrix is utilized to construct a promoted CA matrix for ensemble clustering. We compared our method with 13 state-of-the-art methods across 12 datasets, and the results demonstrated the superiority clustering ability and robustness of the proposed approach. The code is available at https://github.com/xuz2019/SDGCA.

\end{abstract}


\IEEEpeerreviewmaketitle

\section{introduction}
Ensemble clustering (also known as consensus clustering) integrates multiple existing clustering results (referred to as base clusterings) to derive a consensus result that is more accurate and robust than the individual clustering result \cite{1432715}. As ensemble clustering does not require access to the original data features, it is widely applicable in scenarios involving distributed computation, privacy-preserving, and secure data 
 \cite{10.1162/153244303321897735}.

 \begin{figure}
    \centering
    \includegraphics[width=1\linewidth]{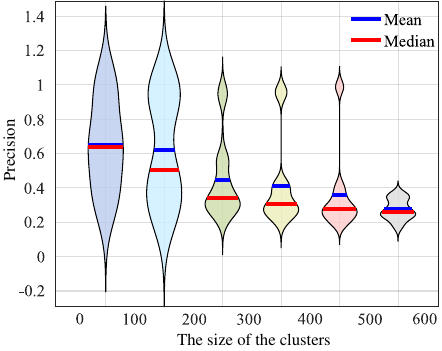}
    \caption{The relationship between the size of clusters and their precision in SPF dataset. The vertical axis represents their mean and median precision. For example, the first violin plot illustrates that clusters with sizes ranging from 0 to 100 have a mean precision and a median precision of 0.65 and 0.64, respectively. It
can be observed smaller cluster sizes generally imply higher precision in both mean and median values.}
    \label{sizeVsacc}
\end{figure}

The co-association (CA) matrix \cite{1432715, berikov2017ensemble,chen2023regularizing}, which records pairwise similarity relationships between data points according to the base clusterings, has garnered significant research attention in recent years. The values within this matrix reflect the confidence level that two data points are assigned to the same cluster. Consequently, the CA matrix is often utilized as an adjacency matrix, undergoing hierarchical clustering or spectral clustering to obtain the final clustering results. In the original method \cite{1432715}, the CA matrix is obtained by simply averaging the adjacency matrices produced by the base clusterings. However, this naive approach does not consider the quality of base clusterings or the clusters in each base clustering. Therefore, scholars have proposed numerous methods to enhance the precision of the CA matrix, which can be categorized into three categories: weighting clusterings, weighting clusters, and weighting objects. For the first category, Vladimir \cite{BERIKOV2017427} proposed a weighting function to evaluate the significance of different base clusterings and theoretically demonstrated that clustering performance improves with the increase in ensemble size. As a representative of methods in weighting clusters, Huang \cite{7932479} introduced locally weighted ensemble clustering, quantifying cluster entropy by the fragmentation of clusters and proposing an inverse relationship between cluster weight and entropy. More references on weighting clusterings and weighting clusters can be found in \cite{zhang2022weighted}. Regarding weighting objectives methods, most methods utilize optimization algorithms to improve the overall CA matrix. For example, Jia \cite{10061157} considered the neighbor relationships with high-confidence samples and proposed a CA matrix self-enhancement model. In \cite{Xu_Li_Duan_2024}, inspired by neighborhood high-order and topological similarity theories, Xu designed an adaptive high-order topological weight method to address the deficiency in handling missing connections in base clustering results. Zhou \cite{10093938} first introduced the idea of utilizing consensus results to refine base clusterings and further improved this idea by employing manifold ranking to diffuse representation across the multiplex graph \cite{10238807}.

\textbf{Motivation}: Despite the significant success of the aforementioned methods, they still suffer from the following two issues.
\begin{enumerate}
    \item They ignore that the quality of the each cluster is highly related to the size of the cluster. As indicated by Fig. \ref{sizeVsacc}. the clusters with smaller size tend to have higher precision, and vice versa
    
    \item They neglect the dissimilarity relationships of the base clusterings. Dissimilarity can evaluate sample pairs that are not directly connected, thereby providing a more detailed reflection of the cluster structure.
\end{enumerate}

To address these issues, this paper proposes a novel method named Similarity and Dissimilarity Guided Co-Association matrix construction (SDGCA), aimed at obtaining an promoted adjacency matrix that fully reflects the relationships between sample points. First, we evaluate the quality of each cluster by normalized ensemble entropy, which effectively compensates for the shortcomings of existing methods in determining cluster weights without considering cluster sizes. Second, random walk is utilized to obtain a dissimilarity relationship matrix for sample pairs. Note that in the CA matrix with a value of 0, all base clusterings do not consider the sample pair to be clustered, but this direct connectivity being 0 does not imply that their indirect connectivity is also 0. Therefore, we need to characterize how likely the sample pair can be separated using a relationship (in this paper, we use dissimilarity). Finally, we find that the above constructed similarity relationships and dissimilarity relationships naturally constitute an adversarial interaction, i.e., a larger (resp. smaller) value in the similarity relation implies a smaller (resp. larger) value in the dissimilarity relation. Through this adversarial prior, we separately learn a promoted similarity matrix and dissimilarity matrix, construct the final adjacency matrix and apply hierarchical clustering for it to obtain the final clustering result.
The main contributions of this paper are summarized as follows:
\begin{itemize}
    \item We observe that the quality of each cluster is highly related to the cluster size, and propose the normalized ensemble entropy to depict the similarity relationship.
    \item A novel perspective for ensemble clustering is proposed by introducing dissimilarity to enhance clustering quality.
    \item We develop a new adversarial learning model to obtain the final adjacency matrix.
    \item Comparative analysis with 13 state-of-the-art methods across 12 datasets demonstrates the superior performance and robustness to hyper-parameters of the proposed method.
\end{itemize}



\section{Related Work}\label{Related Work}
In this section, we first provide a brief introduction to ensemble clustering and the co-association (CA) matrix. Then, we review recent some related ensemble clustering methods. The main notations used in this paper are listed in Table \ref{mainNotion}, other notations will be introduced as they appear.

\begin{table}[htbp]\label{mainNotion}
    \normalsize
    \caption{\textbf{Main notations}}
    \centering
    \begin{tabular}{cp{6.2cm}}
        \toprule
        Notation & Description \\
        \midrule 
        $\mathcal{X}$ & Dataset with $n$ samples \\
        $\Pi$ & Base clustering set with $M$ based clusterings  \\
        $C_{i^m}^m$/$C_i$ & The $i$-th cluster in the $m$-th partition. If there is no superscript, it does not specify which base clustering the cluster belongs to \\
        $N^m$ & Number of clusters in the $m$-th partition \\
        $N_c$ & Number of clusters in the ensemble set \\
        $\mathbf{A}$ & Co-association matrix \\
        $\mathbf{A}_{(:,i)}$/$\mathbf{A}_{(i,:)}$ & The $i$-th column/row elements of matrix \\
        $\mathbf{A}_{ij}$ & Element in the $i$-th row and $j$-th column \\
        $\left\| \mathbf{A}\right\|_\mathrm{F}$ & Frobenius norm of matrix $\mathbf{A}$ \\ 
        $\Tilde{\mathbf{A}}$ & Normalized weighted co-association matrix \\
        $\mathbf{H}$ & High-confidence elements in the CA matrix \\
        $\mathbb{D}$ & Degree matrix, with all off-diagonal elements being zero \\
        $\mathbf{L}$ & Laplacian matrix \\
        $\mathbf{S}$ & Original similarity matrix \\
        $\mathbf{S}^*$ & Learned promoted similarity matrix \\
        $\mathbf{D}$ & Original dissimilarity matrix \\
        $\mathbf{D}^*$ & Learned promoted dissimilarity matrix \\
        $\mathbf{W}$ & Initial adjacency matrix \\
        $\mathbf{W}^*$ & Refined adjacency matrix \\
        
        \bottomrule
    \end{tabular}
\end{table}

\subsection{Ensemble Clustering and CA Matrix}
Let $\mathcal{X}=\{x_1,x_2,\cdots,x_n\}$ be a dataset which contains $N$ samples, and $\Pi = \{\pi^1,\pi^2,\cdots,\pi^M\}$ denotes $M$ base clusterings, where the $m$-th base clustering $\pi^m = \{C^m_1, C^m_2, \cdots, C^m_{N^m}\}$ indicates the clustering partition with $N^m$ clusters. In \cite{1432715}, CA matrix $\mathbf{A} \in \mathbb{R}^{n\times n}$ was proposed:
\begin{equation}\label{CA}
    \mathbf{A}_{ij} = \frac{1}{M}\sum_{k=1}^m \delta_{ij}^m,
\end{equation}
where
\begin{equation*}
\delta _{ij}^{m}=\begin{cases}
	1,&		\mathrm{if}\ \mathrm{Cls}^m\left( x_i \right) =\mathrm{Cls}^{m}\left( x_j \right) \\
	0,&		\mathrm{otherwise}\\
\end{cases},
\end{equation*}
and $\mathrm{Cls}^m$ indicates the cluster in $\pi^m$ that object $x_i$ belongs to. As $\mathbf{A}$ can indicate the pairwise similarities among samples, we typically utilize it to generate the final clustering result. Since Fred and Jain \cite{1432715}, numerous methods for enhancing the CA matrix have been introduced, such as \cite{Jia_Liu_Hou_Zhang_2021, 10173506, 10271692, 10242240, 10.1145/3211872, Li_Qian_Wang_2021, HUANG2023109255, 10070741, 8661522, 9426579}.

\subsection{Locally Weighted Ensemble Clustering}\label{LWCA}
Huang et al. \cite{7932479} proposed using uncertainty to measure the reliability of clusters, thereby assigning different weights to each cluster. Initially, the uncertainty of a cluster is defined as:
\begin{equation}\label{uncertainty_of_cluster}
    H^\Pi(C_i)=-\sum_{j=1}^{N_c}p\Big(C_i,C_j\Big)\log_2p\Big(C_i,C_j\Big),
\end{equation}
where
\begin{equation}\label{p_Ci_Cj}
    p\Big(C_i,C_j\Big)=\frac{\Big|C_i\bigcap C_j\Big|}{|C_i|}.
\end{equation}
Then the ensemble-driven cluster index is proposed to measure the quality of each clustering, i.e.,
\begin{equation}\label{ECI}
    \mathrm{ECI}(C_i)=\mathrm{e}^{-\frac{H^\Pi(C_i)}{\lambda\cdot M}}, 
\end{equation}
where $\lambda$ is a hyper-parameter and $M$ is the number of base clusterings. The ensemble-driven cluster index is used as a weighted adjustment for computing the CA matrix $\mathbf{A}'$.

\subsection{Ensemble Clustering via Co-association Matrix Self-enhancement}
In \cite{10061157}, Jia et al. proposed a CA matrix self-enhancement method. Specifically, it assumes that the LWCA matrix $\mathbf{A}'$ is not sufficiently accurate and is composed of an underlying ideal CA matrix $\mathbf{C}$ and a noise matrix $\mathbf{E}$, such that $\mathbf{A}'=\mathbf{C}+\mathbf{E}$. Then, high-confidence elements $h_{ij}$ in the CA matrix are extracted for propagation. The model is as follows:
\begin{equation}
    \begin{aligned}&\min_{\mathbf{C},\mathbf{E}}\quad\frac{1}{2}\sum_{i,j=1}^{n}h_{ij}\|\mathbf{C}_{i,:}-\mathbf{C}_{j,:}\|_{2}^{2}+\frac{\varphi}{2}\|\mathbf{E}\|_{F}^{2}
    \\
    &\mathrm{s.t.}\quad\mathbf{A}'=\mathbf{C}+\mathbf{E}, \mathbf{C}_{ij}=\mathbf{A}'_{ij}, \forall(i,j)\in\Omega,
    \\
    &\ \ \ \ \ \ \ \ \ \ \ \ \mathbf{C}=\mathbf{C}^\top, \mathbf{0}\leq\mathbf{C}\leq\mathbf{1},\end{aligned}
\end{equation}
where $\Omega$ represents the positions of the high-confidence elements in the CA matrix, $\varphi$ is a hyper-parameter to balance the two loss terms. The authors performed hierarchical clustering on the generated matrix $\mathbf{C}$ to obtain the final clustering results.
\section{Proposed Method}\label{Proposed Method}
In this section, we first introduce how to exploit the similarity relationship by considering the cluster size. Then, we discuss how to characterize the dissimilarity relationship. Afterwards, an adversarial prior is employed to integrate these relationships, constructing an adjacency matrix that yields the final clustering result. Finally, the optimization of this model is discussed.

\subsection{Similarity Relationship Extraction by Normalized Ensemble Entropy}
To characterize the similarity relationships between sample pairs, we approach this from weighting clusters with the following two insights:
\begin{enumerate}
    \item The reliability of a cluster decreases as it receives less support from other clusters.
    \item Smaller cluster sizes lead to tighter connections among sample points within the cluster, resulting in higher precision in pairwise relationships.
\end{enumerate}

For the first perspective, if a cluster $C_1$ suggests that certain sample points should be grouped together, but the majority of other clusters $\{C_2,C_3,\cdots\}$ believe that only a subset of these points should be classified together, it indicates that the cluster $C_1$ exhibits significant uncertainty and thus, lower confidence.

As for the second one, the relationship between cluster size and its precision is depicted in Figure \ref{sizeVsacc}, where we can observe that with larger cluster size, the mean and median precision values decrease from 65\% to 27\%. This occurs because frequently dividing a cluster into smaller groups tends to break weaker connections among them. Consequently, the resulting clusters are composed of sample points that maintain stronger and more stable relationships. 

Based on the above consideration, we propose a novel cluster uncertainty  metric named Normalized Ensemble Entropy (NEE) to estimate the uncertainty of each cluster, which is defined as
\begin{equation}\label{NH}
    \mathrm{NEE}(C_i)=-\frac{\sum_{j=1}^{N_c}p\Big(C_i,C_j\Big)\log_2p\Big(C_i,C_j\Big)}{\log|\pi_{C_i}|},
\end{equation}
where $C_i$ and $C_j$ are the clusters in the ensemble set, $|\pi_{C_i}|$ represents the number of clusters in $\pi_{C_i}$, $p(\cdot,\cdot)$ is defined in Eq. (\ref{p_Ci_Cj}). 

\textbf{Remark 1.} Note that in NEE, we do not directly express it as a function of cluster size, but rather indirectly as the number of clusters in the base clustering. This approach has two advantages: 
\begin{itemize}
    \item Generally, as the number of clusters increases, the size of each cluster naturally decreases.
    \item Even for clusters with a larger size, if they remain intact despite numerous divisions, it is still reasonable to consider that the connections among samples within are very tight.
\end{itemize}

 Furthermore, it can be seen that the definition of cluster uncertainty in LWCA (in Eq. (\ref{uncertainty_of_cluster})) actually represents a special case of our method, when the numbers of clusters in all base clusterings are the same.

Based on NEE, we calculate the weight of each cluster by
\begin{equation}\label{NECI}
    \mathrm{Wei}(C_i)=\text{e}^{-\frac{\text{NEE}\left( C_i \right)}{\lambda \cdot M}}.
\end{equation}
In Eq. (\ref{NECI}), a cluster with higher uncertainty will possess a smaller weight. Then we propose the Normalized Weighted Co-association matrix (NWCA) $\tilde{\mathbf{A}}$ through the weight calculated by Eq. (\ref{NECI}) as
\begin{equation}\label{NWCA}
    \tilde{\mathbf{A}}_{ij}=\frac{1}{M}\sum_{k=1}^m\delta_{ij}^m\cdot\text{Wei}\left(\text{Cls}^m\left(x_i\right)\right).
\end{equation}


Finally, to obtain more accurate information for depicting the similarity relationships between sample pairs, we extract highly credible elements from the NWCA to form the similarity matrix $\mathbf{S}$,
\begin{equation}\label{S}
    \mathbf{S}_{ij}=\begin{cases}
	\tilde{\mathbf{A}}_{ij}&		\mathrm{if}\  \tilde{\mathbf{A}}_{ij}>\eta\\
	0&		\mathrm{else}\\
\end{cases},
\end{equation}
where $\eta$ is a threshold.

The NWCA matrix can act as the initial adjacency matrix for the sample points, denoted as $\mathbf{W}=\tilde{\mathbf{A}}$. In Section \ref{SDGCA}, we will refine $\mathbf{W}$ using the similarity and dissimilarity matrices to obtain the promoted adjacency matrix $\mathbf{W}^*$.

\subsection{Dissimilarity Relationship Extraction by Random Walk}
\begin{figure}
    \centering
    \includegraphics[width=1\linewidth]{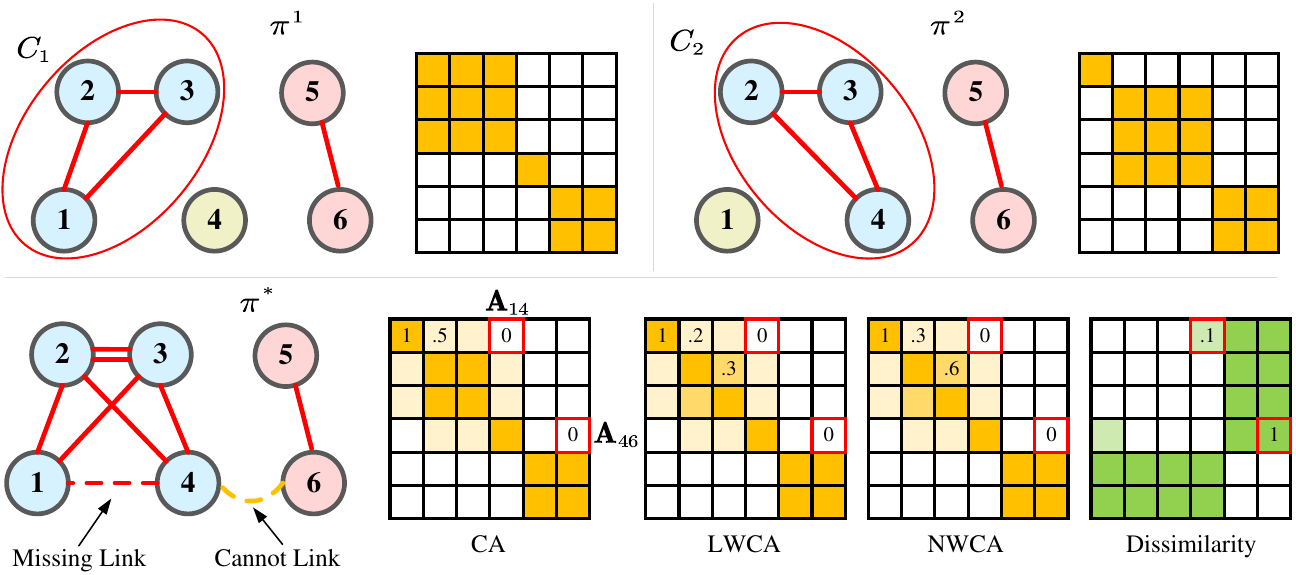}
    \caption{$\pi^1$ and $\pi^2$ represent two base clusterings, dividing six samples, with the corresponding adjacency matrices on the right. $\pi^*$ is the ensemble of $\pi^1$ and $\pi^2$, with the CA, LWCA and NWCA matrices on the right. The far-right side represents the dissimilarity between samples. It is evident in the ensemble that $\{x_1, x_4\}$ are not connected in the same cluster, but intuitively, they should belong to one cluster.}
    \label{diss}
\end{figure}
Although we have recalibrated the similarity relationships between samples, the weighting cluster method is unable to alter the relationships of sample pairs that remain disconnected throughout the entire ensemble pool. As illustrated in Figure \ref{diss}, assume we have two base clusterings, $\pi^1$ and $\pi^2$, with $\pi^*$ representing the result of the ensemble. It can be observed that $x_1$ and $x_4$ are connected to $\{x_2,x_3\}$ in $\pi^1, \pi^2$ respectively. However, as they have never been grouped together, the corresponding element value in the CA matrix of the ensemble result remains zero (i.e., $\mathbf{A}_{14}=0$). Even if we adjust the weights of the clusters, the element values will not change: the corresponding values at these positions in both the LWCA and NWCA matrices are still zero. However, we may think that $x_1$ and $x_4$ likely belong to the same cluster according to $\pi_1$ and $\pi_2$ , indicating a missing link between them. As a comparison, although $\mathbf{A}_{46} = 0$, we do not consider $x_4$ and $x_6$ to be in the same cluster. This is because the clusters $C_1$ containing $x_1$ and $C_2$ containing $x_4$ have high similarity, allowing samples to connect through higher-order relationships, while no higher-oder connections exist between $x_4$ and $x_6$.

\textit{In summary, the zero-valued elements in the CA matrix can exhibit varying confidence levels, which we view as indicators of dissimilarity between samples. Note that, this has been overlooked by all the previous researches.} To capture the dissimilarity relationships, we use random walks to explore direct and higher-order node proximity.

First, we define the direct proximity matrix $\mathbf{P}=[p_{ij}]_{N_c\times N_c}$ of the clusters as
\begin{equation}\label{w_ij}
    p_{ij} = \mathrm{Jaccard}(C_i,C_j).
\end{equation}
where $\mathrm{Jaccard}(C_i,C_j)$ is defined by
\begin{equation}\label{Jaccard}
    \mathrm{Jaccard}\big(C_i,C_j\big)=\frac{|C_i\bigcap C_j|}{|C_i\bigcup C_j|}.
\end{equation}

The normalized proximity matrix $\Tilde{\mathbf{P}}$, which can also be regarded as probability transition matrix is calculated by 
\begin{equation}\label{w_tilde}
    \Tilde{\mathbf{P}} = \mathbb{D}_\mathrm{P}^{-1}\mathbf{P},
\end{equation}
where $\mathbb{D}_{\mathrm{P}}$ is the degree matrix of $\mathbf{P}$.

The higher-order proximity $\tilde{\mathbf{O}}$ is constructed by
\begin{equation}\label{O_tilde}
    \tilde{\mathbf{O}} = \beta^1\tilde{\mathbf{P}}^{\left( 1 \right) \mathrm{T}}\tilde{\mathbf{P}}+   \beta^2\tilde{\mathbf{P}}^{\left( 2 \right) \mathrm{T}}\tilde{\mathbf{P}}+\cdots + \beta^k\tilde{\mathbf{P}}^{\left( k \right) \mathrm{T}}\tilde{\mathbf{P}},
\end{equation}
where $\tilde{\mathbf{P}}^{(i)}=\tilde{\mathbf{P}}^i$, $0\leq \beta \leq 1$ is used to control the weights of different orders. For the robustness of the model, we set $\beta =1$ and $k=20$.

The similarity of clusters $\tilde{\mathbf{R}}=[\tilde{r}_{ij}]_{N_c\times N_c}$ reflecting the higher-order relationships is calculated as
\begin{equation}\label{r_tilde}
\tilde{r}_{ij}=\begin{cases}
	\mathrm{cosine}\left( \tilde{\mathbf{O}}_{\left( :,i \right)},\tilde{\mathbf{O}}_{\left( :,j \right)} \right) ,&		\mathrm{if}\ C_i, C_j\in \pi ^m\\
	0,&		\mathrm{else}\\
\end{cases},
\end{equation}
where $\mathrm{consine}(\cdot, \cdot)$ means the consine similarity between two vectors.

Finally, we get the confidence matrix for the elements in CA matrix with value 0, i.e., dissimilarity matrix of unconnected samples $\mathbf{D}=[\tilde{d}_{ij}]_{n\times n}$ by
\begin{equation}\label{d_tilde}
    \tilde{d}_{ij}=\begin{cases}
	0,&		\text{if} \sum_{u,v}{(1-\tilde{r}_{u,v})} <\tau,\\
	\frac{\sum_{u,v}{(1-\tilde{r}_{u,v})}}{M} &		\text{else},\\
\end{cases}
\end{equation}
where $u=\text{Cls}^m(x_i)$ and $ v=\mathrm{Cls}^m(x_j)$ ($\mathrm{Cls}$ is defined in Eq. (\ref{CA})). $\tau$ is used to filter direct connected samples in CA matrix. As shown in Figure 2, the graph at the bottom right effectively illustrates this dissimilarity. We are 100\% confidence that $x_4$ and $x_6$ are not connected, whereas $x_1$ and $x_4$ have only a 10\% likelihood of being unconnected.

\subsection{Similarity and Dissimilarity Guided Co-association Matrix Construction}\label{SDGCA}
After obtaining the matrices for similar and dissimilar relationships, we aim to construct an adjacency matrix from them and partition it to derive clustering results. First, we obtain two promoted relationship matrices representing the similarity and dissimilarity between sample pairs, denoted as $\mathbf{S}^*$ and $\mathbf{D}^*$, through the following optimization problem.

\begin{equation}\label{trSD_manifold}
\begin{aligned}
    \underset{\mathbf{S}_{ij}^{*},\mathbf{D}_{ij}^{*}}{\min}\ &\sum_{ij}^n{\mathbf{S}_{ij}^{*}\mathbf{D}_{ij}^{*}}
    \\
    +\sum_{i,j=1}^n{\mathbf{H}_{ij}\left\| \mathbf{S}_{\left( :,i \right)}^{*}-\mathbf{S}_{\left( :,j \right)}^{*} \right\| _{2}^{2}}&+ \sum_{i,j=1}^n{\mathbf{H}_{ij}\left\| \mathbf{D}_{\left( :,i \right)}^{*}-\mathbf{D}_{\left( :,j \right)}^{*} \right\| _{2}^{2}}\,\,
\\
\mathrm{s}.\mathrm{t}.\ 0\le \mathbf{S}_{ij}^{*}\le 1,\  0\le \mathbf{D}_{ij}^{*}\le &1, \mathbf{S}^*_{ij}=\mathbf{S}^*_{ji}, \mathbf{D}^*_{ij}=\mathbf{D}^*_{ji},\ \forall (i,j)
\\
\mathbf{S}_{ij}^{*}=\mathbf{S}_{ij},\ \mathbf{D}_{ij}^{*}&=\mathbf{D}_{ij},\ \mathrm{for} \left( i,j \right) \in \Omega .
\end{aligned}
\end{equation}

The motivation for Eq. (\ref{trSD_manifold}) is that, the adversarial relationship exists between elements in $\mathbf{S}^*$ and $\mathbf{D}^*$, i.e., a larger element in $\mathbf{S}^*$ indicates a smaller element in $\mathbf{D}^*$, and vice versa. We express this adversarial prior as the minimization of $\sum_{ij}^n\mathbf{S}_{ij}^{*}\mathbf{D}_{ij}^{*}$ with the constraints $0\leq \mathbf{S}^*_{ij}\leq 1,  0\leq \mathbf{D}^*_{ij}\leq 1, \forall (i,j),$ and $\mathbf{S}^*_{ij}=\mathbf{S}^*_{ij}, \mathbf{D}^*_{ij}=\mathbf{D}^*_{ij}\  \mathrm{for}\  (i,j)\in \Omega$, where $\Omega$ indexes positions of non-zero elements in $\mathbf{S}$ and $\mathbf{D}$, reflecting the assumption that these elements are highly reliable and their respective values should be preserved throughout the learning process. Moreover, we require both $\mathbf{S}^*$ and $\mathbf{D}^*$ to be symmetric by constraining $\mathbf{S}^*_{ij}=\mathbf{S}^*_{ji}, \mathbf{D}^*_{ij}=\mathbf{D}^*_{ji}, \forall i,j$. However, relying solely on this adversarial term can lead to a trivial solution, as $\mathbf{S}^*=\mathbf{S}$ and $\mathbf{D}^*=\mathbf{D}$, from which no valuable information can be derived. Hence, we employ manifold learning to propagate the inherent similarity and dissimilarity. For high-confidence sample pairs in the CA matrix that belong to the same cluster, their similarity relationships with other samples should be as consistent as possible, and the same applies to their dissimilarity relationships, i.e., $\mathbf{H}_{ij}\lVert \mathbf{S}_{\left( :,i \right)}^{*}-\mathbf{S}_{\left( :,j \right)}^{*} \rVert _{2}^{2}$ and $\mathbf{H}_{ij}\lVert \mathbf{D}_{\left( :,i \right)}^{*}-\mathbf{D}_{\left( :,j \right)}^{*} \rVert _{2}^{2}
$ should be minimized, where $\mathbf{H}$ is defined as 

\begin{equation}\label{HC}
    \mathbf{H}_{ij}=\begin{cases}
	\mathbf{A}_{ij},& \mathrm{if}\ \mathbf{A}_{ij}\ge \theta\\
	0,&		 \mathrm{else}\\
\end{cases}.
\end{equation}
If only $\mathbf{H}_{ij}\lVert \mathbf{S}_{\left( :,i \right)}^{*}-\mathbf{S}_{\left( :,j \right)}^{*} \rVert _{2}^{2}$ and $\mathbf{H}_{ij}\lVert \mathbf{D}_{\left( :,i \right)}^{*}-\mathbf{D}_{\left( :,j \right)}^{*} \rVert _{2}^{2}
$ are utilized, the learned $\mathbf{S}^*$ or $\mathbf{D}^*$ will become overly dense, thus losing its significance. Fortunately, the adversarial term $\mathbf{S}_{ij}^{*}\mathbf{D}_{ij}^{*}$ acts as a bridge that will push both $\mathbf{S}^*$ and $\mathbf{D}^*$ to be sparse, where only the reliable information are preserved. 

The optimization problem in Eq. (\ref{trSD_manifold}) can be formulated in matrix form as follows:
\begin{equation}\label{proposed_method}
\begin{aligned}
        \underset{\mathbf{S}^*,\mathbf{D}^*}{\min}\,\,\mathrm{tr}\left( \mathbf{S}^{*\mathrm{T}}  \mathbf{D}^* \right)&+ \mathrm{tr}\left( \mathbf{S}^{*\mathrm{T}}\mathbf{LS}^* \right) + \mathrm{tr}\left( \mathbf{D}^{*\mathrm{T}}\mathbf{LD}^* \right) 
        \\
        \mathrm{s}.\mathrm{t}.\ 0\le \mathbf{S}^*\le 1,0\le &\mathbf{D}^*\le 1,\mathbf{S}^*=\mathbf{S}^{*\mathrm{T}},\mathbf{D}^*=\mathbf{D}^{*\mathrm{T}}
        \\
        \mathcal{P} _{\Omega^\mathbf{S}}\left( \mathbf{S}^* \right)& =\mathbf{S},\mathcal{P} _{\Omega^\mathbf{D}}\left( \mathbf{D}^* \right) =\mathbf{D},
\end{aligned}
\end{equation}
where $\mathbf{L}$ is the Laplacian matrix of $\mathbf{H}$, i.e., $\mathbf{L}=\mathrm{diag}\left( \mathbf{H} \mathbf{1}_n \right) -\mathbf{H}$, $\mathbf{1}_n \in \mathbb{R}^{n}$ is an all-ones column vector. $\mathcal{P}$ is the projection operator, $\mathrm{tr}(\cdot)$ 
 and $(\cdot)^\mathrm{T}$ are used to compute the trace and transpose of a matrix respectively.

Once $\mathbf{S}^*$ and $\mathbf{D}^*$ are acquired, we use them to adjust the initial adjacency matrix $\mathbf{W}$. A natural idea is that for elements where $\mathbf{S}^*_{ij} > \mathbf{D}^*_{ij}$, we should enhance the corresponding value of $\mathbf{W}_{ij}$, and vice versa. Therefore, we update it as follows:
\begin{equation}\label{revise_W}
    \mathbf{W}^*_{ij}=\begin{cases}
    	1-(1-\mathbf{S}^*_{ij}+\mathbf{D}^*_{i,j})(1-\mathbf{W}_{ij}),&		\mathbf{S}^*_{ij}-\mathbf{D}^*_{ij}\ge 0\\
    	(1+\mathbf{S}^*_{ij}-\mathbf{D}^*_{ij})\mathbf{W}_{ij},&		\mathbf{S}^*_{ij}-\mathbf{D}^*_{ij}<0\\
    \end{cases}.
\end{equation}

Finally, we apply hierarchical clustering to $\mathbf{W}^*$ to obtain the final clustering result.

\subsection{Optimization}
To solve Eq. (\ref{proposed_method}), we employ the ADMM (Alternating Direction Method of Multipliers) algorithm \cite{8186925}, which is specifically designed for solving large-scale optimization problems with a decomposable structure. Specifically, ADMM decomposes the original problem into several manageable subproblems and alternates the optimization of each subproblem. Considering the presence of symmetry and range constraints, we introduce intermediate variables $\mathbf{E}$ and $\mathbf{F}$, and reformulate Eq. (\ref{proposed_method}) as follows:
\begin{equation}\label{intermediate}
\begin{aligned}
\underset{\mathbf{S}^*,\mathbf{D}^*,\mathbf{E},\mathbf{F_2}}{\min}\,\,&\mathrm{tr}\left( \mathbf{S}^{*\mathrm{T}}\mathbf{D}^* \right) +  \mathrm{tr}\left( \mathbf{S}^{*\mathrm{T}}\mathbf{LS}^* \right) +   \mathrm{tr}\left( \mathbf{D}^{*\mathrm{T}}\mathbf{LD}^* \right) 
\\
\mathrm{s}.\mathrm{t}.\ \mathbf{E}&=\mathbf{S}^*,\mathbf{F}=\mathbf{D}^*,0\le \mathbf{E}\le 1,0\le \mathbf{F}\le 1,
\\
\mathbf{E}=&\mathbf{E}^{\mathrm{T}},\mathbf{F}=\mathbf{F}^{\mathrm{T}},\mathcal{P} _{\Omega^\mathbf{S}}\left( \mathbf{E} \right) =\mathbf{S},\mathcal{P} _{\Omega^\mathbf{D}}\left( \mathbf{F} \right) =\mathbf{D}.
\end{aligned}
\end{equation}

Define $\Lambda$ and $\Gamma \in \mathbb{R}^{n\times n}$ as the Lagrange multipliers, the augmented Lagrangian function for Eq. (\ref{intermediate}) is
\begin{equation}\label{lagrangian}
    \begin{aligned}
        \mathcal{L} =\mathrm{tr}\left( \mathbf{S}^{*\mathrm{T}}\mathbf{D}^* \right)+  \mathrm{tr}\left( \mathbf{S}^{*\mathrm{T}}\mathbf{LS}^* \right)&+   \mathrm{tr}\left( \mathbf{D}^{*\mathrm{T}}\mathbf{LD}^* \right) 
        \\
        +\mathrm{tr}\left( \Lambda^{\mathrm{T}}\left( \mathbf{S}^*-\mathbf{E} \right) \right) +\frac{\gamma_1}{2}&\left\| \mathbf{S}^*-\mathbf{E} \right\| _{\mathrm{F}}^{2}
        \\
        +\mathrm{tr}\left( \Gamma^{\mathrm{T}}\left(\mathbf{D}^*- \mathbf{F} \right) \right) +\frac{\gamma_2}{2}&\left\| \mathbf{D}^*- \mathbf{F} \right\| _{\mathrm{F}}^{2}
        \\
        \mathrm{s}.\mathrm{t}.\ 0\le \mathbf{E}\le 1, 0\le \mathbf{F}\le 1, \mathbf{E}&=\mathbf{E}^{\mathrm{T}}, \mathbf{F}=\mathbf{F}^{\mathrm{T}}, 
        \\
        \mathcal{P} _{\Omega^\mathbf{S}}\left( \mathbf{E} \right) =\mathbf{S}, \mathcal{P} _{\Omega^\mathbf{D}}(  \mathbf{F} &) =\mathbf{D},
    \end{aligned}
\end{equation}
where $\gamma_1$ and $\gamma_2$ are coefficients associated with quadratic penalty terms. To ensure the stability of convergence, we initialize both $\gamma_1$ and $\gamma_2$ to 1 and increase their values by a factor of 1.1 in each iteration, setting an upper bound of $10^6$ to prevent unbounded growth. We alternately optimize each variable in turn until the algorithm converges.

\begin{enumerate}
\item \textit{Updating} $\mathbf{S^*}$. With other variables fixed, the resulting sub-problem concerning $\mathbf{S^*}$ is formulated as 
\begin{equation}\label{S_subproblem}
    \begin{aligned}
    &\mathbf{S}_{k+1}^{*}=\underset{\mathbf{S}^*}{\mathrm{arg}\min}\,\,\mathrm{tr}\left( \mathbf{S}^{*\mathrm{T}}\mathbf{D}^*_k \right) +  \mathrm{tr}\left( \mathbf{S}^{*\mathrm{T}}\mathbf{LS}^* \right)
    \\
    +&\mathrm{tr}\left( \Lambda^{\mathrm{T}}_k\left( \mathbf{S}^*-\mathbf{E}_k \right) \right) +\frac{\gamma _1}{2}\left\| \mathbf{S}^*-\mathbf{E}_k \right\| _{\mathrm{F}}^{2}.
    \end{aligned}
\end{equation}
where $(\cdot)_k$ represents the value of the variable in the $k$-th iteration. Setting the derivative of this expression with respect to $\mathbf{S^*}$ to zero, we have the updating rule of $\mathbf{S}^*$, i.e., 
\begin{equation}\label{S_solution}
    \mathbf{S}_{k+1}^{*}=\left( 2  \mathbf{L}+\gamma _1\mathbf{I} \right) ^{-1}\left( \gamma _1 \mathbf{E}  _k-\mathbf{D}_{k}^{*}-\Lambda_k \right),
\end{equation}
where $\mathbf{I}\in \mathbb{R}^{n\times n}$ is the identity matrix.

\item \textit{Updating} $\mathbf{E}$. With other variables fixed, the resulting sub-problem concerning $\mathbf{E}$ is formulated as 
\begin{equation}\label{F1_subproblem}
    \begin{aligned}
        \mathbf{E}  _{k+1}=&\underset{\mathbf{E}}{\mathrm{arg}\min}\,\,\mathrm{tr}\left(  \Lambda^{\mathrm{T}}  _k\left( \mathbf{S}_{k}^{*}-\mathbf{E} \right) \right)
        \\
        &+\frac{\gamma_1}{2}\left\| \mathbf{S}_{k}^{*}-\mathbf{E} \right\| _{\mathrm{F}}^{2}
\\
\mathrm{s}.\mathrm{t}. 0\le \mathbf{E}&\le 1, \mathbf{E}=\mathbf{E}^{\mathrm{T}}, \mathcal{P} _{\Omega^\mathbf{S}}\left( \mathbf{E} \right) =\mathbf{S}.
    \end{aligned}
\end{equation}

Although there are three constraints in this subproblem, both the objective function and the constraints are element-wise, we can easily obtain the global solution as
\begin{equation}\label{F1_solution}
\begin{aligned}
        \mathbf{E}_{k+1}=\mathcal{P} _{\Omega^\mathbf{S}}&\left( \max \left( \min \left( \frac{(\mathbf{P}_k+\mathbf{P}_k^\mathrm{T})}{2},1 \right) ,0 \right) \right),
\\
\text{where}\ \ \ \ \ \ ~ \ \ \ \ \ &
\\
&\ \mathbf{P}_k=\mathbf{S}_{k}^{*}+\frac{\Lambda_k}{\gamma _1}.
\end{aligned}
\end{equation}

\item \textit{Updating} $\mathbf{D^*}$. The subproblem of $\mathbf{D}^*$ is 
\begin{equation}\label{D_subproblem}
\begin{aligned}
    &\mathbf{D}_{k+1}^{*}=\underset{\mathbf{S}^*}{\mathrm{arg}\min}\,\,\mathrm{tr}\left( \mathbf{S}_{k}^{*\mathrm{T}}\mathbf{D}^* \right) +  \mathrm{tr}\left( \mathbf{D}^{*\mathrm{T}}\mathbf{LD}^* \right) 
    \\
    +&\mathrm{tr}\left( \Gamma^{\mathrm{T}}_k\left( \mathbf{D}^*-\mathbf{F}_k \right) \right) +\frac{\gamma _2}{2}\left\| \mathbf{D}^*-\mathbf{F}_k \right\| _{\mathrm{F}}^{2}.
\end{aligned}
\end{equation}

Similar to Eq. (\ref{S_solution}), the solution to $\mathbf{D}^*$ is
\begin{equation}\label{D_solution}
    \mathbf{D}^*_{k+1}=\left( 2  \mathbf{L}+\gamma _2\mathbf{I} \right) ^{-1}\left( \gamma _2\mathbf{F}_k-\mathbf{S}_k-\Gamma _k \right) .
\end{equation}

\item \textit{Updating} $\mathbf{F}$. We solve the subproblem for $\mathbf{F}$
\begin{equation}\label{F2_subproblem}
    \begin{aligned}
        \mathbf{F} _{k+1}=&\underset{\mathbf{E}}{\mathrm{arg}\min}\,\,\mathrm{tr}\left( \Gamma^{\mathrm{T}}_k\left( \mathbf{D}_{k}^{*}-\mathbf{F} \right) \right)
        \\
        &+\frac{\gamma _2}{2}\left\| \mathbf{D}_{k}^{*}-\mathbf{F} \right\| _{\mathrm{F}}^{2}
        \\
        \mathrm{s}.\mathrm{t}.\ 0\le \mathbf{F}&\le 1, \mathbf{F}=\mathbf{F}^{\mathrm{T}}, \mathcal{P} _{\Omega^\mathbf{D}}\left( \mathbf{F} \right) =\mathbf{D}. 
    \end{aligned}
\end{equation}

We use the same method as solving $\mathbf{E}$ to obtain
\begin{equation}\label{F2_solution}
    \begin{aligned}
        \mathbf{F} _{k+1}=\mathcal{P} _{\Omega^\mathbf{D}}&\left( \max \left( \min \left( \frac{\mathbf{Q}_k+\mathbf{Q}_k^\mathrm{T}}{2},1 \right) ,0 \right) \right),
        \\
        \text{where}\ \ \ \ \ \ \ \ ~ \ \ \ \ \ &
        \\
        &\mathbf{Q}_k=\mathbf{D}_{k}^{*}+\frac{\Gamma _k}{\gamma _2}.
    \end{aligned}   
\end{equation}

\item Updating $\Lambda$ and $\Gamma$. Finally, we update the Lagrange multipliers as follows:
\begin{equation} \label{Y_solution}
    \begin{aligned}
        \Lambda_{k+1}&=\Lambda_k+\gamma _1\left( \mathbf{S}_{k}^{*}- \mathbf{E}_k \right)
        \\
        \Gamma_{k+1}&= \Gamma_k+\gamma _2\left( \mathbf{D}_{k}^{*}-\mathbf{F}_k \right).
    \end{aligned}
\end{equation}
\end{enumerate}

The above optimization process is summarized in Algorithm 1.

\subsection{Computational Complexity Analysis}
The time complexity of the proposed method primarily lies in the construction of the NWCA matrix, the dissimilarity matrix $ \mathbf{D} $, and the Algorithm 1. For NWCA matrix, the computation is $ \mathcal{O}(nN_c+n^2N_c) $, where $n$ is the number of samples, $N_c$ is the number of clusters in the ensemble set. In calculation of $ \mathbf{D} $, the time complexity of computing Eqs. (\ref{w_ij}-\ref{d_tilde}) is $ \mathcal{O}(n(N_c)^2+(N_c)^2+(N_c)^{2.37}+(N_c)^3+(n^2N_c+n(N_c)^2+n^2)) $. In Algorithm 1, the main time cost lies in matrix multiplication and matrix inversion. Since $\left( 2  \mathbf{L}+\gamma _1\mathbf{I} \right) ^{-1}$ only needs to be computed once and saved, the latest matrix multiplication techniques \cite{10353208} enable us to achieve a time complexity of $n^{2.37}$ in each iteration.

\subsection{Convergence Analysis}
Because problem (\ref{intermediate}) is non-convex, we provide a convergence proof for it under some mild assumptions. To simplify notation, we define $\mathbf{Z}=\left( \mathbf{S}^*,\mathbf{D}^*,\mathbf{E},\mathbf{F}, \Gamma,\Lambda \right) $ and $\mathcal{L}(\cdot)$ is the augmented lagrangian function.

\textbf{Theorem 1}. Assume $\left\{ \left( \mathbf{Z}_k,\Gamma _k, \Lambda _k \right) \right\} $ is a sequence in $k$-th iteration calculated by ADMM. If $\left\{ \left(  \Gamma_k, \Lambda_k \right) \right\}, \mathrm{tr}(\mathbf{S}^{*\mathrm{T}}\mathbf{D}^*)$ are bounded and 
\begin{equation}\label{assum}
    \sum_{k=0}^\infty\left(\|\Lambda_{k+1}-\Lambda_k\|_\mathrm{F}^2+\|\Gamma_{k+1}-\Gamma_k\|_\mathrm{F}^2\right)<\infty.
\end{equation}
Then sequence $\left\{ \left( \mathbf{Z}_k,\Lambda _k, \Gamma _k \right) \right\} $ will converge and any accumulation point of $\mathbf{Z}_k$ satisfies the KKT condition.

The proof of \textbf{Theorem 1} is shown in the Appendix \ref{Convergence}.

\begin{algorithm}[!htbp]\label{a1}
    \caption{Similarity and Dissimilarity Guided Co-association Matrix Construction}
    \label{alg:AOA}
    \renewcommand{\algorithmicrequire}{\textbf{Input:}}
    \renewcommand{\algorithmicensure}{\textbf{Output:}}
    \newcommand{\algorithmicinitialize}{\textbf{Initialization:}}
    \newcommand{\INITIALIZE}{\item[\algorithmicinitialize]}
    
    \begin{algorithmic}[1]
        \REQUIRE Multiple base clustering results $\Pi$.
        \INITIALIZE Penalty parameter $(\gamma_1)_k,(\gamma_2)_k = 1$, growth factor $\rho=1.1$, upper bound $\gamma_{\mathbf{max}}=10^6$,  tolerance $\epsilon =1e-3$, $k=0$, $\mathbf{S}_{k}^{*}=\mathbf{D}_{k}^{*}= \mathbf{E} _k=\mathbf{F}_k=\Lambda_k=\Gamma _k=\mathbf{0}\in \mathbb{R}^{n\times n}$.
        \ENSURE The revised affinity matrix $\mathbf{W^*}$.

        \STATE Calculate CA matrix by Eq. (\ref{CA}), and calculate its high-confidence matrix by Eq. (\ref{HC}).
        \STATE  Obtain $\mathbf{W}=\tilde{\mathbf{A}}$ and $\mathbf{S}$ by Eqs. (\ref{NWCA}), (\ref{S}), and obtain $\mathbf{D}$ by Eq (\ref{d_tilde}).
        \STATE Calculate Laplacian matrix $\mathbf{L}$ with $\mathbf{L}=\mathrm{diag}\left( \mathbf{H} \mathbf{1}_N \right) -\mathbf{H}$.
        \WHILE {not converged}
            \STATE Update $\mathbf{S}^*_{k+1}$ by Eq. (\ref{S_solution}).
            \STATE Update $\mathbf{E}_{k+1}$ by Eq. (\ref{F1_solution}).
            \STATE Update $\mathbf{D}^*_{k+1}$ by Eq. (\ref{D_solution}).
            \STATE Update $\mathbf{F}_{k+1}$ by Eq. (\ref{F2_solution}).
            \STATE Update $\Lambda_{k+1}$ and $\Gamma_{k+1}$ by Eq. (\ref{Y_solution}).
            \STATE $(\gamma_1)_{k+1}=\min (\rho\cdot (\gamma_1)_k, \gamma_{max})$.
            \STATE$(\gamma_2)_{k+1}=\min (\rho\cdot (\gamma_2)_k, \gamma_{max})$.
            \IF{$\sigma_k(\mathbf{S}^*),\sigma_k(\mathbf{D}^*),\sigma_k(\mathbf{E}),\sigma_k(\mathbf{F}) \le \epsilon$}
            \STATE Break (With $\sigma_k(\mathbf{S}^*)=(\|\mathbf{S}^*_{k+1}-\mathbf{S}^*_k\|_\mathrm{F}^2)/\|\mathbf{S}^*_k\|_\mathrm{F}^2$).
            \ENDIF
        \ENDWHILE
        \STATE Revise $\mathbf{W}$ by Eq. (\ref{revise_W}).
        \RETURN Revised affinity matrix $\mathbf{W}^*$.
    \end{algorithmic}
\end{algorithm}

\begin{table}[ht]
    \normalsize
    \caption{Details of different datasets}
    \label{datasets}
    \centering
    \begin{tabular}{ccccc}
        \toprule
        No. & Dataset & \#Instance & \#Feature & \#Class \\
        \midrule 
        D1 & Ecoli \cite{ecoli_39} & 336 & 8 & 8 \\
        D2 & GLIOMA & 50 & 4434 & 4\\
        D3 & Aggregation \cite{gionis2007clustering} & 788 & 2 & 7 \\
        D4 & MF & 2,000 & 649 & 10 \\
        D5 & IS & 2,310 & 19 & 7 \\
        D6 & MNIST & 5,000 & 784 & 10 \\
        D7 & Texture & 5,500 & 40 & 11\\
        D8 & SPF & 1941 & 27 & 7 \\
        D9 & ODR & 5,620 & 64 & 10 \\
        D10 & LS & 6,435  & 36 & 6\\
        D11 & ISOLET \cite{isolet_54} & 7,797 & 617 & 26\\
        D12 & USPS & 11,000 & 256 & 10\\
        \bottomrule
    \end{tabular}
\end{table}

\begin{table*}[htbp]
    \normalsize
    \caption{The clustering performances of different algorithms are measured by NMI, with the best and second-best performances highlighted in bold and underlined.}
    \label{NMI}
    \centering
    \scalebox{0.92}{
    \begin{tabular}{|c|c|c|c|c|c|c|c|c|c|c|c|c|}
        \toprule
        Method  & D1 &  D2 & D3 &   D4   &   D5   & D6  & D7 & D8 & D9 & D10 & D11 & D12 \\
        \midrule 
$k$-means (best)				& $\underset{\pm 0.000 }{0.669 }$ & $\underset{\pm 0.000 }{0.653 }$ & $\underset{\pm 0.000 }{0.878 }$ & $\underset{\pm 0.000 }{0.643 }$ & $\underset{\pm 0.000 }{0.671 }$ & $\underset{\pm 0.000 }{0.580 }$ & $\underset{\pm 0.000 }{0.698 }$ & $\underset{\pm 0.000 }{0.180 }$ & $\underset{\pm 0.000 }{0.784 }$ & $\underset{\pm 0.000 }{0.611 }$ & $\underset{\pm 0.000 }{0.749 }$ & $\underset{\pm 0.000 }{0.578 }$ \\
\midrule
$k$-means (average)				& $\underset{\pm 0.041 }{0.579 }$ & $\underset{\pm 0.066 }{0.491 }$ & $\underset{\pm 0.073 }{0.743 }$ & $\underset{\pm 0.071 }{0.599 }$ & $\underset{\pm 0.044 }{0.600 }$ & $\underset{\pm 0.055 }{0.537 }$ & $\underset{\pm 0.064 }{0.655 }$ & $\underset{\pm 0.029 }{0.157 }$ & $\underset{\pm 0.077 }{0.692 }$ & $\underset{\pm 0.050 }{0.507 }$ & $\underset{\pm 0.051 }{0.702 }$ & $\underset{\pm 0.055 }{0.535 }$ \\
\midrule
\midrule
EAC				& $\underset{\pm 0.028 }{0.632 }$ & $\underset{\pm 0.015 }{0.501 }$ & $\underset{\pm 0.041 }{0.926 }$ & $\underset{\pm 0.023 }{0.628 }$ & $\underset{\pm 0.017 }{0.610 }$ & $\underset{\pm 0.016 }{0.625 }$ & $\underset{\pm 0.022 }{0.705 }$ & $\underset{\pm 0.021 }{0.118 }$ & $\underset{\pm 0.017 }{0.803 }$ & $\underset{\pm 0.064 }{0.581 }$ & $\underset{\pm 0.008 }{0.757 }$ & $\underset{\pm 0.013 }{0.578 }$ \\
\midrule
PTA-CL				& $\underset{\pm 0.022 }{0.557 }$ & $\underset{\pm 0.090 }{0.499 }$ & $\underset{\pm 0.084 }{0.784 }$ & $\underset{\pm 0.029 }{0.652 }$ & $\underset{\pm 0.036 }{0.631 }$ & $\underset{\pm 0.025 }{0.620 }$ & $\underset{\pm 0.027 }{0.729 }$ & $\underset{\pm 0.032 }{0.141 }$ & $\underset{\pm 0.018 }{0.825 }$ & $\underset{\pm 0.038 }{0.572 }$ & $\underset{\pm 0.013 }{0.737 }$ & $\underset{\pm 0.022 }{0.574 }$ \\
\midrule
PTA-SL				& $\underset{\pm 0.043 }{0.586 }$ & $\underset{\pm 0.046 }{0.506 }$ & $\underset{\pm 0.037 }{0.849 }$ & $\underset{\pm 0.053 }{0.625 }$ & $\underset{\pm 0.034 }{0.586 }$ & $\underset{\pm 0.124 }{0.200 }$ & $\underset{\pm 0.068 }{0.558 }$ & $\underset{\pm 0.040 }{0.140 }$ & $\underset{\pm 0.057 }{0.700 }$ & $\underset{\pm 0.091 }{0.024 }$ & $\underset{\pm 0.022 }{0.701 }$ & $\underset{\pm 0.158 }{0.125 }$ \\
\midrule
CEAM				& $\underset{\pm 0.031 }{0.542 }$ & $\underset{\pm 0.084 }{0.473 }$ & $\underset{\pm 0.023 }{0.877 }$ & $\underset{\pm 0.029 }{0.569 }$ & $\underset{\pm 0.036 }{0.599 }$ & $\underset{\pm 0.025 }{0.592 }$ & $\underset{\pm 0.042 }{0.591 }$ & $\underset{\pm 0.025 }{0.117 }$ & $\underset{\pm 0.052 }{0.744 }$ & $\underset{\pm 0.083 }{0.475 }$ & $\underset{\pm 0.015 }{0.713 }$ & $\underset{\pm 0.038 }{0.514 }$ \\
\midrule
LWEA				& $\underset{\pm 0.021 }{0.629 }$ & $\underset{\pm 0.025 }{0.501 }$ & $\underset{\pm 0.046 }{0.941 }$ & $\underset{\pm 0.017 }{0.677 }$ & $\underset{\pm 0.020 }{0.642 }$ & $\underset{\pm 0.015 }{0.662 }$ & $\underset{\pm 0.019 }{0.789 }$ & $\underset{\pm 0.028 }{0.151 }$ & $\underset{\pm 0.011 }{0.837 }$ & $\underset{\pm 0.024 }{0.636 }$ & $\underset{\pm 0.009 }{0.767 }$ & $\underset{\pm 0.016 }{0.658 }$ \\
\midrule
RSEC-Z				& $\underset{\pm 0.016 }{0.575 }$ & $\underset{\pm 0.017 }{0.493 }$ & $\underset{\pm 0.023 }{0.840 }$ & $\underset{\pm 0.022 }{0.630 }$ & $\underset{\pm 0.011 }{0.611 }$ & $\underset{\pm 0.029 }{0.539 }$ & $\underset{\pm 0.016 }{0.681 }$ & $\underset{\pm 0.013 }{0.079 }$ & $\underset{\pm 0.023 }{0.787 }$ & $\underset{\pm 0.052 }{0.555 }$ & $\underset{\pm 0.007 }{0.741 }$ & $\underset{\pm 0.021 }{0.617 }$ \\
\midrule
RSEC-H				& $\underset{\pm 0.031 }{0.543 }$ & $\underset{\pm 0.076 }{0.494 }$ & $\underset{\pm 0.055 }{0.754 }$ & $\underset{\pm 0.029 }{0.618 }$ & $\underset{\pm 0.021 }{0.599 }$ & $\underset{\pm 0.032 }{0.548 }$ & $\underset{\pm 0.027 }{0.666 }$ & $\underset{\pm 0.021 }{0.097 }$ & $\underset{\pm 0.026 }{0.790 }$ & $\underset{\pm 0.053 }{0.493 }$ & $\underset{\pm 0.009 }{0.716 }$ & $\underset{\pm 0.025 }{0.603 }$ \\
\midrule
ECPCS-M				& $\underset{\pm 0.021 }{0.653 }$ & $\underset{\pm 0.029 }{0.489 }$ & $\underset{\pm 0.030 }{0.939 }$ & $\underset{\pm 0.024 }{0.668 }$ & $\underset{\pm 0.013 }{0.653 }$ & $\underset{\pm 0.015 }{0.643 }$ & $\underset{\pm 0.019 }{0.731 }$ & $\underset{\pm 0.014 }{0.140 }$ & $\underset{\pm 0.008 }{0.828 }$ & $\underset{\pm 0.010 }{0.626 }$ & $\underset{\pm 0.003 }{0.756 }$ & $\underset{\pm 0.013 }{0.623 }$ \\
\midrule
TRCE				& $\underset{\pm 0.051 }{0.662 }$ & $\underset{\pm 0.027 }{0.506 }$ & $\underset{\pm 0.010 }{0.983 }$ & $\underset{\pm 0.023 }{0.660 }$ & $\underset{\pm 0.016 }{0.623 }$ & $\underset{\pm 0.010 }{0.648 }$ & $\underset{\pm 0.024 }{0.743 }$ & $\underset{\pm 0.033 }{0.133 }$ & $\underset{\pm 0.012 }{0.828 }$ & $\underset{\pm 0.036 }{0.648 }$ & $\underset{\pm 0.005 }{0.756 }$ & $\underset{\pm 0.031 }{0.639 }$ \\
\midrule
CESHL				& $\underset{\pm 0.044 }{0.656 }$ & $\underset{\pm 0.011 }{0.506 }$ & $\underset{\pm 0.005 }{0.982 }$ & $\underset{\pm 0.029 }{0.661 }$ & $\underset{\pm 0.017 }{0.644 }$ & $\underset{\pm 0.010 }{0.643 }$ & $\underset{\pm 0.027 }{0.759 }$ & $\underset{\pm 0.023 }{0.148 }$ & $\underset{\pm 0.007 }{0.833 }$ & $\underset{\pm 0.025 }{0.654 }$ & $\underset{\pm 0.006 }{0.759 }$ & $\underset{\pm 0.028 }{0.626 }$ \\
\midrule
SCCABG				& $\underset{\pm 0.040 }{0.642 }$ & $\underset{\pm 0.078 }{0.484 }$ & $\underset{\pm 0.025 }{0.978 }$ & $\underset{\pm 0.027 }{0.635 }$ & $\underset{\pm 0.027 }{0.624 }$ & $\underset{\pm 0.036 }{0.633 }$ & $\underset{\pm 0.068 }{0.654 }$ & $\underset{\pm 0.017 }{0.125 }$ & $\underset{\pm 0.031 }{0.817 }$ & $\underset{\pm 0.088 }{0.543 }$ & $\underset{\pm 0.009 }{0.748 }$ & $\underset{\pm 0.069 }{0.504 }$ \\
\midrule
ECCMS				& $\underset{\pm 0.042 }{\underline{0.695 }}$ & $\underset{\pm 0.018 }{\underline{0.507 }}$ & $\underset{\pm 0.003 }{\underline{0.984 }}$ & $\underset{\pm 0.015 }{\underline{0.699 }}$ & $\underset{\pm 0.029 }{\underline{0.656 }}$ & $\underset{\pm 0.016 }{\underline{0.668 }}$ & $\underset{\pm 0.019 }{\underline{0.806 }}$ & $\underset{\pm 0.020 }{\underline{0.184 }}$ & $\underset{\pm 0.016 }{\underline{0.841 }}$ & $\underset{\pm 0.035 }{\underline{0.663 }}$ & $\underset{\pm 0.007 }{\underline{0.769 }}$ & $\underset{\pm 0.012 }{0.662 }$ \\
\midrule
AWEC-H				& $\underset{\pm 0.019 }{0.677 }$ & $\underset{\pm 0.022 }{0.501 }$ & $\underset{\pm 0.031 }{0.965 }$ & $\underset{\pm 0.024 }{0.687 }$ & $\underset{\pm 0.019 }{0.638 }$ & $\underset{\pm 0.008 }{0.657 }$ & $\underset{\pm 0.031 }{0.785 }$ & $\underset{\pm 0.031 }{0.152 }$ & $\underset{\pm 0.023 }{0.833 }$ & $\underset{\pm 0.025 }{0.653 }$ & $\underset{\pm 0.013 }{0.743 }$ & $\underset{\pm 0.025 }{\underline{0.668 }}$ \\
\midrule
\midrule
SDGCA				& $\underset{\pm 0.028}{\mathbf{0.721}}$ & $\underset{\pm 0.036}{\mathbf{0.538}}$ & $\underset{\pm 0.002}{\mathbf{0.985}}$ & $\underset{\pm 0.016}{\mathbf{0.700}}$ & $\underset{\pm 0.023}{\mathbf{0.697}}$ & $\underset{\pm 0.017}{\mathbf{0.678}}$ & $\underset{\pm 0.016}{\mathbf{0.814}}$ & $\underset{\pm 0.026}{\mathbf{0.199}}$ & $\underset{\pm 0.013}{\mathbf{0.856}}$ & $\underset{\pm 0.021}{\mathbf{0.685}}$ & $\underset{\pm 0.008}{\mathbf{0.771}}$ & $\underset{\pm 0.013}{\mathbf{0.678}}$ \\
        \bottomrule
    \end{tabular}}
\end{table*}

\begin{table*}[htbp]
    \normalsize
    \caption{Ablation experiments on NMI metric, where “w/o” indicates the removal of the corresponding component from the model.}
    \label{Ablation_NMI}
    \centering
    \scalebox{0.93}{
    \begin{tabular}{|ccccccccccccc|}
        \toprule
        Method  & D1 &  D2 & D3 &   D4   &   D5   & D6  & D7 & D8 & D9 & D10 & D11 & D12\\
        \midrule 
        SDGCA    & $0.721$	& $\mathbf{0.538}$ & $\mathbf{0.985}$ & $\mathbf{0.700}$	& $\mathbf{0.697}$ &	$\mathbf{0.678}$	&$ \mathbf{0.814}$	& $\mathbf{0.199}$                   & $\mathbf{0.856} $& $\mathbf{0.685}$ & $\mathbf{0.771}$ & $\mathbf{0.678}$ \\
        \midrule 
        Only $\mathbf{S}^*$ & $\mathbf{0.724}$ & $0.516$ & $0.984$ & $0.380$ & $0.014$ & $0.004$ & $0.004$ & $0.076$ & $0.015$ & $0.538$ & $0.035$  &  $0.029$\\
        NWCA & $0.617$	& $0.520$ &	$0.972$ & $0.695$	& $0.690$ & $0.676$ & $\mathbf{0.814 }$ &	$0.181$ & $0.856$ &	$0.660$ & $0.769$ & $\mathbf{0.678}$ \\
        w/o $\mathrm{tr}(\mathbf{S}^{*\mathrm{T}}\mathbf{LS}^*)$  & $0.636$ & $0.513$ & $0.980$ & $0.697$ & $0.694$ & $0.678$ & $\mathbf{0.814}$ & $0.185$ & $0.852$ & $0.669$ & $0.769$ & $\mathbf{0.678}$\\
        w/o $\mathrm{tr}(\mathbf{D}^{*\mathrm{T}}\mathbf{LD}^*)$  & $0.713$ & $0.527$ & $0.982$ & $0.681$ & $0.690$ & $0.678$ & $\mathbf{0.814}$ & $0.197$ & $0.850$ & $0.672$ & $0.770$ & $\mathbf{0.678}$ \\
        \parbox{2.39cm}{w/o $\mathrm{tr}(\mathbf{S}^{*\mathrm{T}}\mathbf{LS}^*) \\ + \mathrm{tr}(\mathbf{D}^{*\mathrm{T}}\mathbf{LD}^*)$} & $0.636$ & $0.513$ & $0.980$ & $0.697$ & $0.694$ & $0.678$ & $\mathbf{0.814}$ & $0.185$ & $0.852$ & $0.650$ & $0.769$ & $\mathbf{0.678}$\\

        \bottomrule
    \end{tabular}}
\end{table*}

\begin{table*}[htbp]
    \normalsize
    \caption{The clustering performances of different algorithms are measured by ARI, with the best and second-best performances highlighted in bold and underlined.}
    \label{ARI}
    \centering
    \scalebox{0.92}{
    \begin{tabular}{|c|c|c|c|c|c|c|c|c|c|c|c|c|}
        \toprule
        Method  & D1 &  D2 & D3 &   D4   &   D5   & D6  & D7 & D8 & D9 & D10 & D11 & D12 \\
        \midrule 
       $k$-means (best)				& $\underset{\pm 0.000 }{0.524 }$ & $\underset{\pm 0.000 }{0.480 }$ & $\underset{\pm 0.000 }{0.808 }$ & $\underset{\pm 0.000 }{0.451 }$ & $\underset{\pm 0.000 }{0.481 }$ & $\underset{\pm 0.000 }{0.315 }$ & $\underset{\pm 0.000 }{0.530 }$ & $\underset{\pm 0.000 }{0.056 }$ & $\underset{\pm 0.000 }{0.745 }$ & $\underset{\pm 0.000 }{0.479 }$ & $\underset{\pm 0.000 }{0.511 }$ & $\underset{\pm 0.000 }{0.293 }$ \\
\midrule
$k$-means (average)				& $\underset{\pm 0.120 }{0.396 }$ & $\underset{\pm 0.062 }{0.326 }$ & $\underset{\pm 0.171 }{0.463 }$ & $\underset{\pm 0.080 }{0.374 }$ & $\underset{\pm 0.111 }{0.344 }$ & $\underset{\pm 0.064 }{0.260 }$ & $\underset{\pm 0.088 }{0.373 }$ & $\underset{\pm 0.013 }{0.044 }$ & $\underset{\pm 0.146 }{0.428 }$ & $\underset{\pm 0.100 }{0.235 }$ & $\underset{\pm 0.068 }{0.412 }$ & $\underset{\pm 0.066 }{0.215 }$ \\
\midrule
\midrule
EAC				& $\underset{\pm 0.072 }{0.467 }$ & $\underset{\pm 0.020 }{0.329 }$ & $\underset{\pm 0.091 }{0.860 }$ & $\underset{\pm 0.026 }{0.470 }$ & $\underset{\pm 0.030 }{0.492 }$ & $\underset{\pm 0.014 }{0.485 }$ & $\underset{\pm 0.037 }{0.558 }$ & $\underset{\pm 0.011 }{0.055 }$ & $\underset{\pm 0.037 }{0.707 }$ & $\underset{\pm 0.102 }{0.491 }$ & $\underset{\pm 0.018 }{0.529 }$ & $\underset{\pm 0.025 }{0.398 }$ \\
\midrule
PTA-CL				& $\underset{\pm 0.038 }{0.329 }$ & $\underset{\pm 0.107 }{0.349 }$ & $\underset{\pm 0.154 }{0.599 }$ & $\underset{\pm 0.036 }{0.495 }$ & $\underset{\pm 0.049 }{0.507 }$ & $\underset{\pm 0.038 }{0.505 }$ & $\underset{\pm 0.042 }{0.597 }$ & $\underset{\pm 0.019 }{0.077 }$ & $\underset{\pm 0.035 }{0.757 }$ & $\underset{\pm 0.059 }{0.462 }$ & $\underset{\pm 0.027 }{0.497 }$ & $\underset{\pm 0.034 }{0.396 }$ \\
\midrule
PTA-SL				& $\underset{\pm 0.121 }{0.457 }$ & $\underset{\pm 0.058 }{\underline{0.353 }}$ & $\underset{\pm 0.056 }{0.759 }$ & $\underset{\pm 0.081 }{0.405 }$ & $\underset{\pm 0.059 }{0.347 }$ & $\underset{\pm 0.031 }{0.042 }$ & $\underset{\pm 0.084 }{0.207 }$ & $\underset{\pm 0.026 }{0.068 }$ & $\underset{\pm 0.121 }{0.431 }$ & $\underset{\pm 0.050 }{0.012 }$ & $\underset{\pm 0.063 }{0.272 }$ & $\underset{\pm 0.049 }{0.036 }$ \\
\midrule
CEAM				& $\underset{\pm 0.041 }{0.334 }$ & $\underset{\pm 0.075 }{0.312 }$ & $\underset{\pm 0.051 }{0.745 }$ & $\underset{\pm 0.035 }{0.415 }$ & $\underset{\pm 0.039 }{0.481 }$ & $\underset{\pm 0.030 }{0.435 }$ & $\underset{\pm 0.054 }{0.411 }$ & $\underset{\pm 0.015 }{0.069 }$ & $\underset{\pm 0.080 }{0.625 }$ & $\underset{\pm 0.083 }{0.334 }$ & $\underset{\pm 0.029 }{0.450 }$ & $\underset{\pm 0.040 }{0.340 }$ \\
\midrule
LWEA				& $\underset{\pm 0.052 }{0.451 }$ & $\underset{\pm 0.037 }{0.335 }$ & $\underset{\pm 0.107 }{0.893 }$ & $\underset{\pm 0.024 }{0.527 }$ & $\underset{\pm 0.021 }{0.517 }$ & $\underset{\pm 0.026 }{\underline{0.553 }}$ & $\underset{\pm 0.028 }{0.702 }$ & $\underset{\pm 0.024 }{0.082 }$ & $\underset{\pm 0.023 }{0.787 }$ & $\underset{\pm 0.053 }{0.579 }$ & $\underset{\pm 0.016 }{\underline{0.567 }}$ & $\underset{\pm 0.034 }{0.513 }$ \\
\midrule
RSEC-Z				& $\underset{\pm 0.027 }{0.371 }$ & $\underset{\pm 0.023 }{0.323 }$ & $\underset{\pm 0.053 }{0.689 }$ & $\underset{\pm 0.027 }{0.506 }$ & $\underset{\pm 0.011 }{0.500 }$ & $\underset{\pm 0.028 }{0.400 }$ & $\underset{\pm 0.027 }{0.535 }$ & $\underset{\pm 0.014 }{0.049 }$ & $\underset{\pm 0.049 }{0.718 }$ & $\underset{\pm 0.071 }{0.418 }$ & $\underset{\pm 0.018 }{0.515 }$ & $\underset{\pm 0.023 }{0.454 }$ \\
\midrule
RSEC-H				& $\underset{\pm 0.056 }{0.335 }$ & $\underset{\pm 0.081 }{0.352 }$ & $\underset{\pm 0.089 }{0.568 }$ & $\underset{\pm 0.038 }{0.482 }$ & $\underset{\pm 0.021 }{0.489 }$ & $\underset{\pm 0.030 }{0.395 }$ & $\underset{\pm 0.034 }{0.504 }$ & $\underset{\pm 0.026 }{0.057 }$ & $\underset{\pm 0.062 }{0.712 }$ & $\underset{\pm 0.061 }{0.328 }$ & $\underset{\pm 0.025 }{0.467 }$ & $\underset{\pm 0.029 }{0.436 }$ \\
\midrule
ECPCS-M				& $\underset{\pm 0.020 }{0.526 }$ & $\underset{\pm 0.027 }{0.332 }$ & $\underset{\pm 0.064 }{0.913 }$ & $\underset{\pm 0.040 }{0.535 }$ & $\underset{\pm 0.015 }{\underline{0.534 }}$ & $\underset{\pm 0.021 }{0.528 }$ & $\underset{\pm 0.037 }{0.589 }$ & $\underset{\pm 0.010 }{0.065 }$ & $\underset{\pm 0.011 }{0.777 }$ & $\underset{\pm 0.029 }{0.532 }$ & $\underset{\pm 0.010 }{0.523 }$ & $\underset{\pm 0.015 }{0.467 }$ \\
\midrule
TRCE				& $\underset{\pm 0.111 }{0.623 }$ & $\underset{\pm 0.028 }{0.298 }$ & $\underset{\pm 0.018 }{0.986 }$ & $\underset{\pm 0.028 }{0.506 }$ & $\underset{\pm 0.027 }{0.508 }$ & $\underset{\pm 0.009 }{0.480 }$ & $\underset{\pm 0.057 }{0.581 }$ & $\underset{\pm 0.027 }{0.065 }$ & $\underset{\pm 0.021 }{0.758 }$ & $\underset{\pm 0.088 }{0.549 }$ & $\underset{\pm 0.022 }{0.502 }$ & $\underset{\pm 0.043 }{0.462 }$ \\
\midrule
CESHL				& $\underset{\pm 0.147 }{0.577 }$ & $\underset{\pm 0.010 }{0.335 }$ & $\underset{\pm 0.005 }{0.986 }$ & $\underset{\pm 0.041 }{0.504 }$ & $\underset{\pm 0.011 }{0.531 }$ & $\underset{\pm 0.010 }{0.490 }$ & $\underset{\pm 0.060 }{0.620 }$ & $\underset{\pm 0.014 }{0.067 }$ & $\underset{\pm 0.012 }{0.773 }$ & $\underset{\pm 0.062 }{0.582 }$ & $\underset{\pm 0.014 }{0.531 }$ & $\underset{\pm 0.041 }{0.439 }$ \\
\midrule
SCCABG				& $\underset{\pm 0.116 }{0.514 }$ & $\underset{\pm 0.077 }{0.313 }$ & $\underset{\pm 0.044 }{0.976 }$ & $\underset{\pm 0.060 }{0.445 }$ & $\underset{\pm 0.060 }{0.481 }$ & $\underset{\pm 0.089 }{0.429 }$ & $\underset{\pm 0.130 }{0.325 }$ & $\underset{\pm 0.017 }{0.043 }$ & $\underset{\pm 0.089 }{0.726 }$ & $\underset{\pm 0.122 }{0.356 }$ & $\underset{\pm 0.025 }{0.487 }$ & $\underset{\pm 0.089 }{0.229 }$ \\
\midrule
ECCMS				& $\underset{\pm 0.095 }{0.697 }$ & $\underset{\pm 0.034 }{0.346 }$ & $\underset{\pm 0.004 }{\underline{0.989 }}$ & $\underset{\pm 0.029 }{0.552 }$ & $\underset{\pm 0.035 }{0.517 }$ & $\underset{\pm 0.011 }{0.528 }$ & $\underset{\pm 0.044 }{\underline{0.712 }}$ & $\underset{\pm 0.020 }{\underline{0.102 }}$ & $\underset{\pm 0.029 }{\underline{0.795 }}$ & $\underset{\pm 0.060 }{\underline{0.616 }}$ & $\underset{\pm 0.016 }{0.555 }$ & $\underset{\pm 0.023 }{0.495 }$ \\
\midrule
AWEC-H				& $\underset{\pm 0.026 }{\underline{0.730 }}$ & $\underset{\pm 0.036 }{0.341 }$ & $\underset{\pm 0.055 }{0.954 }$ & $\underset{\pm 0.032 }{\underline{0.556 }}$ & $\underset{\pm 0.024 }{0.494 }$ & $\underset{\pm 0.028 }{0.517 }$ & $\underset{\pm 0.061 }{0.665 }$ & $\underset{\pm 0.027 }{0.101 }$ & $\underset{\pm 0.043 }{0.764 }$ & $\underset{\pm 0.047 }{0.588 }$ & $\underset{\pm 0.032 }{0.497 }$ & $\underset{\pm 0.043 }{\underline{0.535 }}$ \\

\midrule
\midrule
SDGCA				& $\underset{\pm 0.061}{\mathbf{0.748}}$ & $\underset{\pm 0.031}{\mathbf{0.384}}$ & $\underset{\pm 0.001}{\mathbf{0.990}}$ & $\underset{\pm 0.029}{\mathbf{0.565}}$ & $\underset{\pm 0.033}{\mathbf{0.548}}$ & $\underset{\pm 0.039}{\mathbf{0.571}}$ & $\underset{\pm 0.035}{\mathbf{0.739}}$ & $\underset{\pm 0.017}{\mathbf{0.109}}$ & $\underset{\pm 0.027}{\mathbf{0.826}}$ & $\underset{\pm 0.029}{\mathbf{0.658}}$ & $\underset{\pm 0.013}{\mathbf{0.577}}$ & $\underset{\pm 0.028}{\mathbf{0.552}}$ \\
        \bottomrule
    \end{tabular}}
\end{table*}

\begin{table*}[htbp]
    \normalsize
    \caption{Ablation experiments on ARI metric, where “w/o” indicates the removal of the corresponding component from the model.}
    \label{Ablation_ARI}
    \centering
    \scalebox{0.93}{
    \begin{tabular}{|ccccccccccccc|}
        \toprule
        Method  & D1 &  D2 & D3 &   D4   &   D5   & D6  & D7 & D8 & D9 & D10 & D11 & D12\\
        \midrule 
        SDGCA   & $0.748$	& $\mathbf{0.384}$       &	$\mathbf{0.990}$         &	$\mathbf{0.565}$         &	$\mathbf{0.548}$         &	$\mathbf{0.571}$	        & $\mathbf{0.739}$  & $\mathbf{0.109}$       & $\mathbf{0.826}$ & $\mathbf{0.658}$ & $\mathbf{0.577}$ & $\mathbf{0.552}$\\
        \midrule 
        Only $\mathbf{S}^*$  & $\mathbf{0.749}$ & $0.363$ & $0.989$ & $0.197$ & $0.001$ & $0.000$ & $0.000$ & $0.019$ & $0.001$ & $0.385$ & $0.001$ & $0.001$\\
        NWCA & $0.452$	& $0.365$	& $0.963$	& $0.555$	& $0.542$	& $0.567$	& $\mathbf{0.739}$	& $0.104$	& $0.825$	& $0.622$ & $0.572$ & $\mathbf{0.552}$\\
        w/o $\mathrm{tr}(\mathbf{S}^{*\mathrm{T}}\mathbf{LS}^*)$  & $0.487$ & $0.356$ & $0.977$ & $0.560$ & $0.544$ & $0.569$ & $\mathbf{0.739}$ & $0.106$ & $0.817$ & $0.645$ & $0.574$ & $\mathbf{0.552}$\\
        w/o $\mathrm{tr}(\mathbf{D}^{*\mathrm{T}}\mathbf{LD}^*)$   & $0.719$ & $0.371$ & $0.984$ & $0.496$ & $0.546$ & $0.570$ & $\mathbf{0.739}$ & $0.108$ & $0.817$ & $0.620$ & $0.576$ & $\mathbf{0.552}$\\
        \parbox{2.39cm}{w/o $\mathrm{tr}(\mathbf{S}^{*\mathrm{T}}\mathbf{LS}^*) \\ + \mathrm{tr}(\mathbf{D}^{*\mathrm{T}}\mathbf{LD}^*)$} & $0.487$ & $0.356$ & $0.977$ & $0.560$ & $0.544$ & $0.569$ & $\mathbf{0.739}$ & $0.106$ & $0.817$ & $0.598$ & $0.574$ & $\mathbf{0.552}$\\
        \bottomrule
    \end{tabular}}
\end{table*}

\begin{table*}[htbp]
    \normalsize
    \caption{The clustering performances of different algorithms are measured by F-score, with the best and second-best performances highlighted in bold and underlined.}
    \label{F_score}
    \centering
    \scalebox{0.92}{
    \begin{tabular}{|c|c|c|c|c|c|c|c|c|c|c|c|c|}
        \toprule
        Method  & D1 &  D2 & D3 &   D4   &   D5   & D6  & D7 & D8 & D9 & D10 & D11 & D12\\
        \midrule 
        $k$-means (best)				& $\underset{\pm 0.000 }{0.625 }$ & $\underset{\pm 0.000 }{0.653 }$ & $\underset{\pm 0.000 }{0.847 }$ & $\underset{\pm 0.000 }{0.492 }$ & $\underset{\pm 0.000 }{0.581 }$ & $\underset{\pm 0.000 }{0.351 }$ & $\underset{\pm 0.000 }{0.561 }$ & $\underset{\pm 0.000 }{0.160 }$ & $\underset{\pm 0.000 }{0.770 }$ & $\underset{\pm 0.000 }{0.555 }$ & $\underset{\pm 0.000 }{0.532 }$ & $\underset{\pm 0.000 }{0.327 }$ \\
\midrule
$k$-means (average)				& $\underset{\pm 0.127 }{0.504 }$ & $\underset{\pm 0.060 }{0.498 }$ & $\underset{\pm 0.174 }{0.528 }$ & $\underset{\pm 0.072 }{0.423 }$ & $\underset{\pm 0.117 }{0.400 }$ & $\underset{\pm 0.076 }{0.301 }$ & $\underset{\pm 0.086 }{0.410 }$ & $\underset{\pm 0.064 }{0.146 }$ & $\underset{\pm 0.144 }{0.462 }$ & $\underset{\pm 0.121 }{0.291 }$ & $\underset{\pm 0.065 }{0.432 }$ & $\underset{\pm 0.079 }{0.249 }$ \\
\midrule
\midrule
EAC				& $\underset{\pm 0.061 }{0.577 }$ & $\underset{\pm 0.022 }{0.515 }$ & $\underset{\pm 0.074 }{0.886 }$ & $\underset{\pm 0.023 }{0.528 }$ & $\underset{\pm 0.025 }{0.569 }$ & $\underset{\pm 0.012 }{0.542 }$ & $\underset{\pm 0.032 }{0.601 }$ & $\underset{\pm 0.011 }{0.264 }$ & $\underset{\pm 0.033 }{0.739 }$ & $\underset{\pm 0.084 }{0.590 }$ & $\underset{\pm 0.017 }{0.549 }$ & $\underset{\pm 0.022 }{0.466 }$ \\
\midrule
PTA-CL				& $\underset{\pm 0.029 }{0.454 }$ & $\underset{\pm 0.063 }{0.526 }$ & $\underset{\pm 0.116 }{0.682 }$ & $\underset{\pm 0.031 }{0.550 }$ & $\underset{\pm 0.038 }{0.585 }$ & $\underset{\pm 0.034 }{0.557 }$ & $\underset{\pm 0.037 }{0.635 }$ & $\underset{\pm 0.023 }{0.249 }$ & $\underset{\pm 0.031 }{0.782 }$ & $\underset{\pm 0.049 }{0.562 }$ & $\underset{\pm 0.025 }{0.518 }$ & $\underset{\pm 0.030 }{0.461 }$ \\
\midrule
PTA-SL				& $\underset{\pm 0.100 }{0.580 }$ & $\underset{\pm 0.040 }{0.530 }$ & $\underset{\pm 0.042 }{0.815 }$ & $\underset{\pm 0.066 }{0.483 }$ & $\underset{\pm 0.041 }{0.474 }$ & $\underset{\pm 0.023 }{0.212 }$ & $\underset{\pm 0.066 }{0.325 }$ & $\underset{\pm 0.028 }{0.294 }$ & $\underset{\pm 0.098 }{0.512 }$ & $\underset{\pm 0.029 }{0.325 }$ & $\underset{\pm 0.056 }{0.317 }$ & $\underset{\pm 0.036 }{0.208 }$ \\
\midrule
CEAM				& $\underset{\pm 0.036 }{0.452 }$ & $\underset{\pm 0.058 }{0.502 }$ & $\underset{\pm 0.041 }{0.794 }$ & $\underset{\pm 0.031 }{0.478 }$ & $\underset{\pm 0.032 }{0.560 }$ & $\underset{\pm 0.027 }{0.497 }$ & $\underset{\pm 0.047 }{0.471 }$ & $\underset{\pm 0.015 }{0.258 }$ & $\underset{\pm 0.071 }{0.665 }$ & $\underset{\pm 0.067 }{0.459 }$ & $\underset{\pm 0.028 }{0.473 }$ & $\underset{\pm 0.035 }{0.410 }$ \\
\midrule
LWEA				& $\underset{\pm 0.045 }{0.563 }$ & $\underset{\pm 0.030 }{0.515 }$ & $\underset{\pm 0.088 }{0.913 }$ & $\underset{\pm 0.022 }{0.579 }$ & $\underset{\pm 0.017 }{0.595 }$ & $\underset{\pm 0.023 }{\underline{0.601 }}$ & $\underset{\pm 0.025 }{0.729 }$ & $\underset{\pm 0.021 }{0.278 }$ & $\underset{\pm 0.021 }{0.809 }$ & $\underset{\pm 0.042 }{0.664 }$ & $\underset{\pm 0.015 }{\underline{0.586 }}$ & $\underset{\pm 0.029 }{0.568 }$ \\
\midrule
RSEC-Z				& $\underset{\pm 0.023 }{0.481 }$ & $\underset{\pm 0.020 }{0.504 }$ & $\underset{\pm 0.043 }{0.749 }$ & $\underset{\pm 0.024 }{0.557 }$ & $\underset{\pm 0.008 }{0.573 }$ & $\underset{\pm 0.026 }{0.463 }$ & $\underset{\pm 0.024 }{0.579 }$ & $\underset{\pm 0.019 }{0.231 }$ & $\underset{\pm 0.043 }{0.747 }$ & $\underset{\pm 0.057 }{0.526 }$ & $\underset{\pm 0.017 }{0.534 }$ & $\underset{\pm 0.020 }{0.518 }$ \\
\midrule
RSEC-H				& $\underset{\pm 0.049 }{0.464 }$ & $\underset{\pm 0.054 }{\underline{0.533 }}$ & $\underset{\pm 0.069 }{0.652 }$ & $\underset{\pm 0.033 }{0.536 }$ & $\underset{\pm 0.018 }{0.564 }$ & $\underset{\pm 0.028 }{0.461 }$ & $\underset{\pm 0.030 }{0.552 }$ & $\underset{\pm 0.023 }{0.243 }$ & $\underset{\pm 0.054 }{0.743 }$ & $\underset{\pm 0.045 }{0.462 }$ & $\underset{\pm 0.023 }{0.489 }$ & $\underset{\pm 0.025 }{0.501 }$ \\
\midrule
ECPCS-M				& $\underset{\pm 0.014 }{0.625 }$ & $\underset{\pm 0.023 }{0.515 }$ & $\underset{\pm 0.052 }{0.930 }$ & $\underset{\pm 0.035 }{0.584 }$ & $\underset{\pm 0.011 }{\underline{0.609 }}$ & $\underset{\pm 0.019 }{0.577 }$ & $\underset{\pm 0.033 }{0.629 }$ & $\underset{\pm 0.008 }{0.275 }$ & $\underset{\pm 0.010 }{0.800 }$ & $\underset{\pm 0.023 }{0.618 }$ & $\underset{\pm 0.010 }{0.543 }$ & $\underset{\pm 0.013 }{0.525 }$ \\
\midrule
TRCE				& $\underset{\pm 0.081 }{0.724 }$ & $\underset{\pm 0.033 }{0.466 }$ & $\underset{\pm 0.014 }{0.989 }$ & $\underset{\pm 0.024 }{0.560 }$ & $\underset{\pm 0.022 }{0.583 }$ & $\underset{\pm 0.008 }{0.540 }$ & $\underset{\pm 0.049 }{0.624 }$ & $\underset{\pm 0.027 }{0.284 }$ & $\underset{\pm 0.018 }{0.784 }$ & $\underset{\pm 0.060 }{0.647 }$ & $\underset{\pm 0.020 }{0.523 }$ & $\underset{\pm 0.038 }{0.525 }$ \\
\midrule
CESHL				& $\underset{\pm 0.129 }{0.674 }$ & $\underset{\pm 0.010 }{0.520 }$ & $\underset{\pm 0.004 }{0.989 }$ & $\underset{\pm 0.035 }{0.559 }$ & $\underset{\pm 0.011 }{0.605 }$ & $\underset{\pm 0.009 }{0.547 }$ & $\underset{\pm 0.052 }{0.658 }$ & $\underset{\pm 0.026 }{0.300 }$ & $\underset{\pm 0.011 }{0.796 }$ & $\underset{\pm 0.045 }{0.668 }$ & $\underset{\pm 0.014 }{0.551 }$ & $\underset{\pm 0.035 }{0.505 }$ \\
\midrule
SCCABG				& $\underset{\pm 0.101 }{0.617 }$ & $\underset{\pm 0.038 }{0.509 }$ & $\underset{\pm 0.034 }{0.982 }$ & $\underset{\pm 0.047 }{0.511 }$ & $\underset{\pm 0.047 }{0.567 }$ & $\underset{\pm 0.069 }{0.502 }$ & $\underset{\pm 0.104 }{0.417 }$ & $\underset{\pm 0.031 }{0.295 }$ & $\underset{\pm 0.074 }{0.757 }$ & $\underset{\pm 0.079 }{0.527 }$ & $\underset{\pm 0.023 }{0.510 }$ & $\underset{\pm 0.067 }{0.347 }$ \\
\midrule
ECCMS				& $\underset{\pm 0.066 }{0.786 }$ & $\underset{\pm 0.030 }{0.529 }$ & $\underset{\pm 0.003 }{\underline{0.991 }}$ & $\underset{\pm 0.025 }{0.602 }$ & $\underset{\pm 0.028 }{0.596 }$ & $\underset{\pm 0.010 }{0.585 }$ & $\underset{\pm 0.039 }{\underline{0.740 }}$ & $\underset{\pm 0.016 }{ \underline{0.306} }$ & $\underset{\pm 0.026 }{\underline{0.816 }}$ & $\underset{\pm 0.044 }{\underline{0.700 }}$ & $\underset{\pm 0.015 }{0.576 }$ & $\underset{\pm 0.019 }{0.564 }$ \\
\midrule
AWEC-H				& $\underset{\pm 0.018 }{\underline{0.802 }}$ & $\underset{\pm 0.024 }{0.517 }$ & $\underset{\pm 0.043 }{0.964 }$ & $\underset{\pm 0.028 }{\underline{0.602 }}$ & $\underset{\pm 0.019 }{0.578 }$ & $\underset{\pm 0.023 }{0.574 }$ & $\underset{\pm 0.054 }{0.688 }$ & $\underset{\pm 0.020 }{0.303 }$ & $\underset{\pm 0.038 }{0.789 }$ & $\underset{\pm 0.037 }{0.675 }$ & $\underset{\pm 0.029 }{0.526 }$ & $\underset{\pm 0.038 }{\underline{0.586 }}$ \\

\midrule
\midrule
SDGCA				& $\underset{\pm 0.047}{\mathbf{0.819}}$ & $\underset{\pm 0.031}{\mathbf{0.568}}$ & $\underset{\pm 0.001}{\mathbf{0.992}}$ & $\underset{\pm 0.025}{\mathbf{0.612}}$ & $\underset{\pm 0.025}{\mathbf{0.624}}$ & $\underset{\pm 0.033}{\mathbf{0.618}}$ & $\underset{\pm 0.031}{\mathbf{0.763}}$ & $\underset{\pm 0.024}{\mathbf{0.340}}$ & $\underset{\pm 0.024}{\mathbf{0.843}}$ & $\underset{\pm 0.021}{\mathbf{0.731}}$ & $\underset{\pm 0.012}{\mathbf{0.595}}$ & $\underset{\pm 0.024}{\mathbf{0.602}}$ \\
        \bottomrule
    \end{tabular}}
\end{table*}

\begin{table*}[htbp]
    \normalsize
    \caption{Ablation experiments on F-score metric, where “w/o” indicates the removal of the corresponding component from the model.}
    \label{Ablation_F}
    \centering
    \scalebox{0.93}{
    \begin{tabular}{|ccccccccccccc|}
        \toprule
        Method  & D1 &  D2 & D3 &   D4   &   D5   & D6  & D7 & D8 & D9 & D10 & D11 & D12\\
        \midrule 
 SDGCA & $0.819$       & $\mathbf{0.568}$       &	$\mathbf{0.992}$    &	$\mathbf{0.612}$	        & $\mathbf{0.624}$       &	$\mathbf{0.618}$         &  $\mathbf{0.763}$  & $\mathbf{0.340}$       & $\mathbf{0.843}$  & $\mathbf{0.731}$ & $\mathbf{0.595}$ & $\mathbf{0.602}$ \\
        \midrule 
        Only $\mathbf{S}^*$   & $\mathbf{0.820}$ & $0.557$ & $0.991$ & $0.330$ & $0.249$ & $0.181$ & $0.166$ & $0.261$ & $0.182$ & $0.550$ & $0.075$ & $0.133$\\
        NWCA & $0.565$	& $0.546$	& $0.971$	& $0.604$	& $0.618$	& $0.614$	& $\mathbf{0.763}$	& $0.300$ & 	$0.843$ &	$0.698$ & $0.592$ & $\mathbf{0.602}$\\
        w/o $\mathrm{tr}(\mathbf{S}^{*\mathrm{T}}\mathbf{LS}^*)$  & $0.594$ & $0.544$ & $0.982$ & $0.609$ & $0.621$ & $0.617$ & $\mathbf{0.763}$ & $0.310$ & $0.836$ & $0.720$ & $0.593$ & $\mathbf{0.602}$\\
        w/o $\mathrm{tr}(\mathbf{D}^{*\mathrm{T}}\mathbf{LD}^*)$   & $0.795$ & $0.556$ & $0.988$ & $0.560$ & $0.622$ & $0.616$ & $\mathbf{0.763}$ & $0.337$ & $0.836$ & $0.705$ & $0.594$ & $\mathbf{0.602}$\\
        \parbox{2.39cm}{w/o $\mathrm{tr}(\mathbf{S}^{*\mathrm{T}}\mathbf{LS}^*) \\ + \mathrm{tr}(\mathbf{D}^{*\mathrm{T}}\mathbf{LD}^*)$}  & $0.594$ & $0.544$ & $0.982$ & $0.609$ & $0.621$ & $0.617$ & $\mathbf{0.763}$ & $0.310$ & $0.836$ & $0.681$ & $0.593$  & $\mathbf{0.602}$\\
        \bottomrule
    \end{tabular}}
\end{table*}

\section{Experiment}\label{Experiment}
To validate the effectiveness of our method, in this section, we compared the proposed SDGCA with 13 state-of-the-art methods across 12 real-world datasets.

\subsection{Datasets and Performance Evaluation Metrics}
We collected datasets of varying sizes, features, and numbers of clustering clusters from multiple sources for experiments, namely Ecoli, GLIOMA, Aggregation, MF, IS (image segmentation), MNIST, Texture, SPF (steel plates faults),  ODR, LS (letter recognition), ISOLET, and USPS. The detailed information about those datasets are summarized in Table \ref{datasets}. Following \cite{10061157}, we used the $k$-means algorithm with the value of $k$ ranging from $[2, \sqrt{n}]$ to randomly generate 100 candidate base clusterings for each dataset, where $n$ is the number of samples. Then, we randomly selected 20 base clusterings for ensemble clustering each time, repeated this process 20 times, and reported the mean and variance of the results.

For performance evaluation, we employed three widely used metrics: NMI (Normalized Mutual Information), ARI (Adjusted Rand Index), and F-score. The value range of ARI is from -1 to 1, while both NMI and F-score range from 0 to 1. For all three metrics, higher values indicate that the clustering results are more consistent with the ground truth.

\begin{figure*}
    \centering
    \subfigure[Ecoli]{\label{para_Ecoli}
    \includegraphics[width=0.23\linewidth]{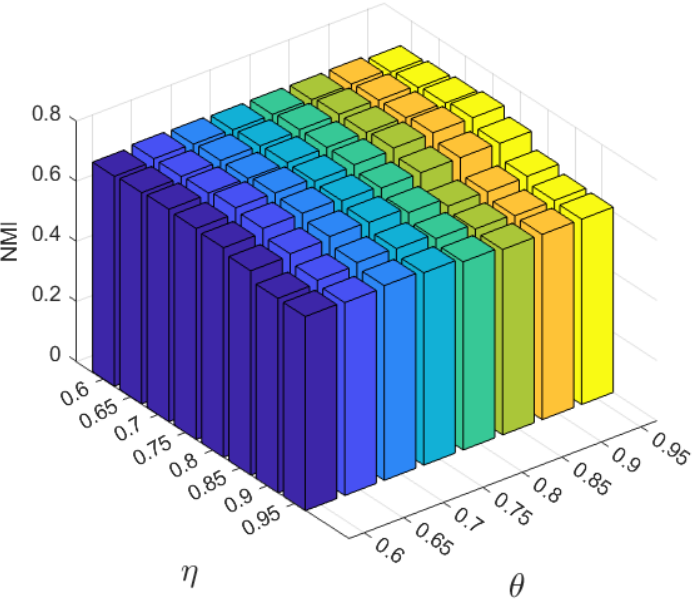}}
    \hspace{0.001\linewidth}
    \subfigure[GLIOMA]{\label{para_GLIOMA}
    \includegraphics[width=0.23\linewidth]{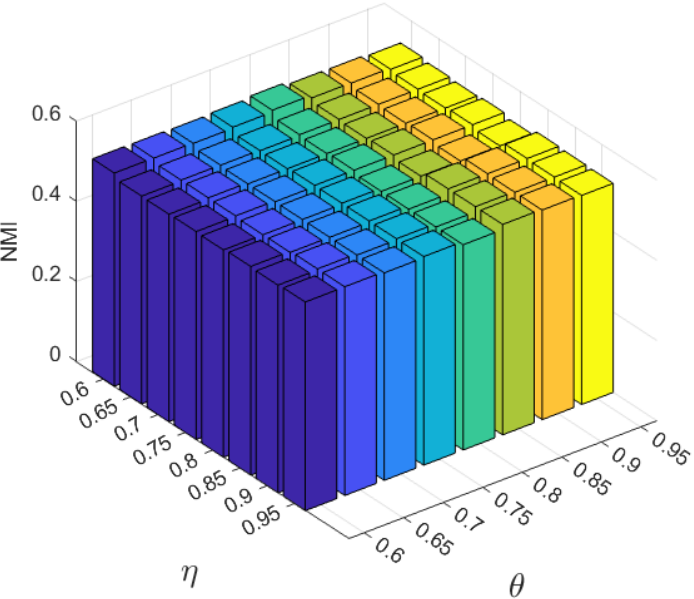}}
    \hspace{0.001\linewidth}
    \subfigure[Aggregation]{\label{para_Aggregation}
    \includegraphics[width=0.23\linewidth]{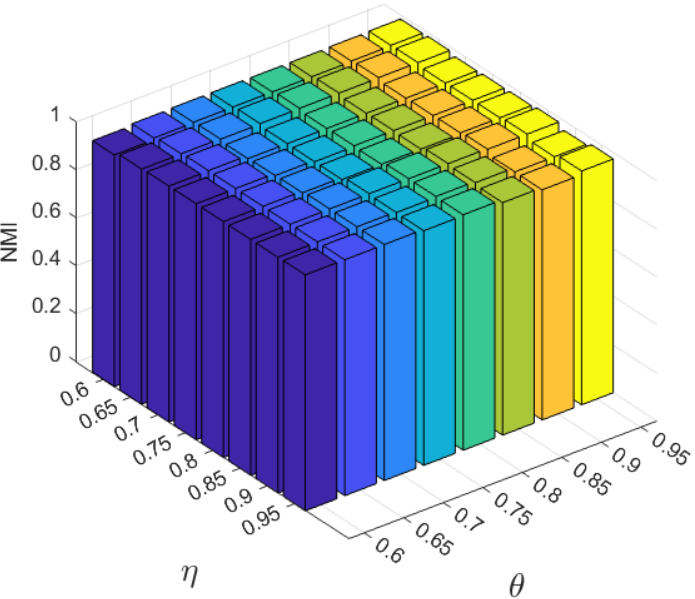}}
    \hspace{0.001\linewidth}
    \subfigure[MF]{\label{para_MF}
    \includegraphics[width=0.23\linewidth]{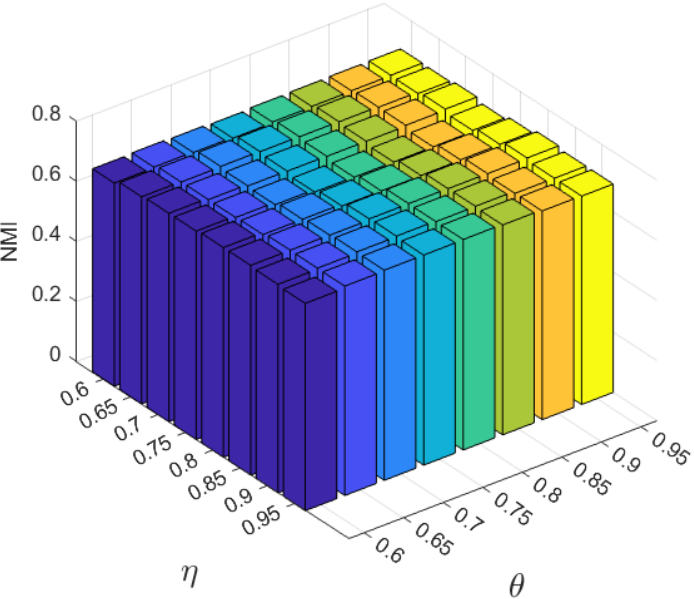}}
        \hspace{0.001\linewidth}
    \subfigure[IS]{\label{para_IS}
    \includegraphics[width=0.23\linewidth]{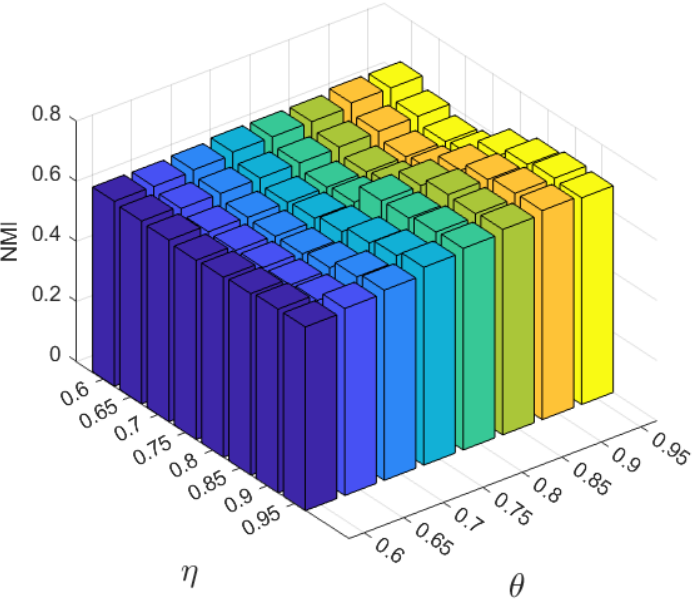}}
        \hspace{0.001\linewidth}
    \subfigure[MNIST]{\label{para_MNIST}
    \includegraphics[width=0.23\linewidth]{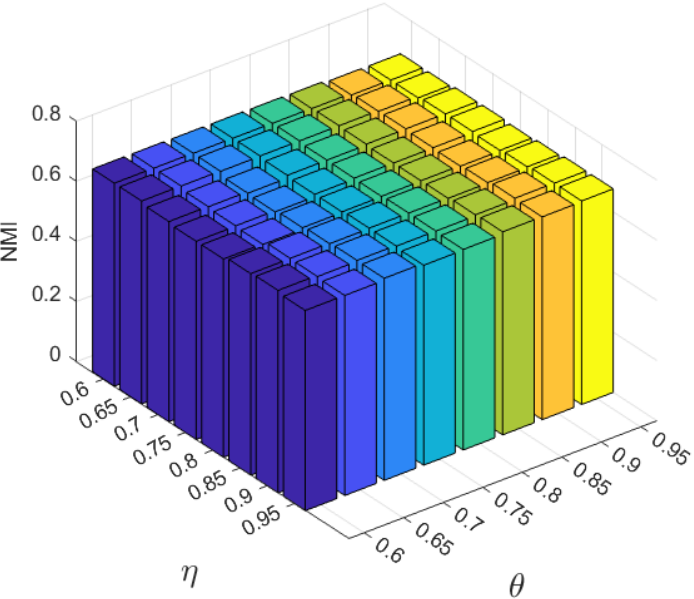}}
        \hspace{0.001\linewidth}
    \subfigure[Texture]{\label{para_Texture}
    \includegraphics[width=0.23\linewidth]{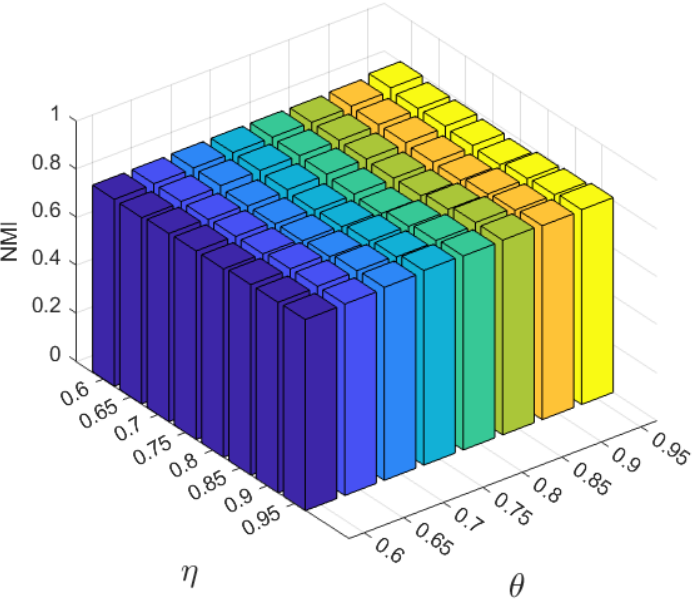}}
        \hspace{0.001\linewidth}
    \subfigure[SPF]{\label{para_SPF}
    \includegraphics[width=0.23\linewidth]{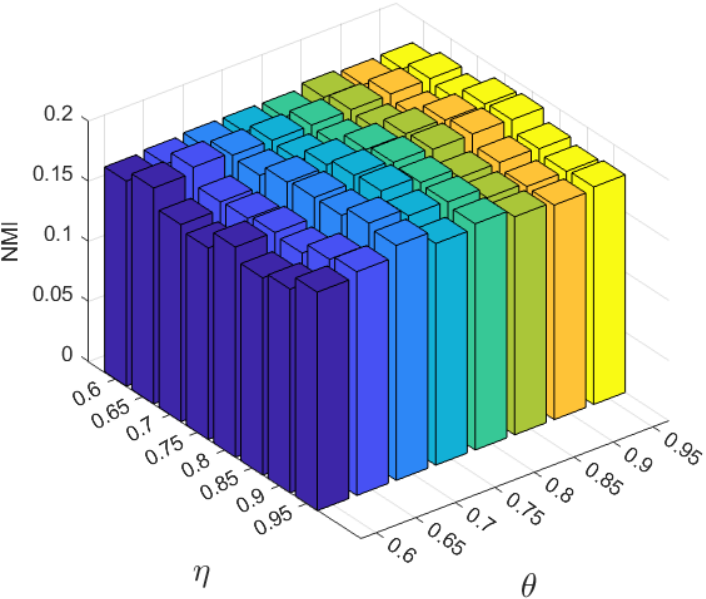}}
            \hspace{0.001\linewidth}
    \subfigure[ODR]{\label{para_ODR}
    \includegraphics[width=0.23\linewidth]{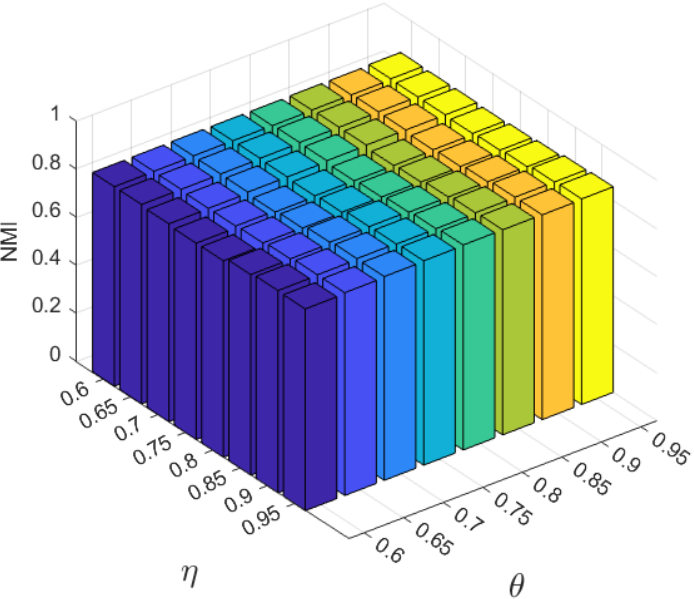}}
            \hspace{0.001\linewidth}
    \subfigure[LS]{\label{para_LS}
    \includegraphics[width=0.23\linewidth]{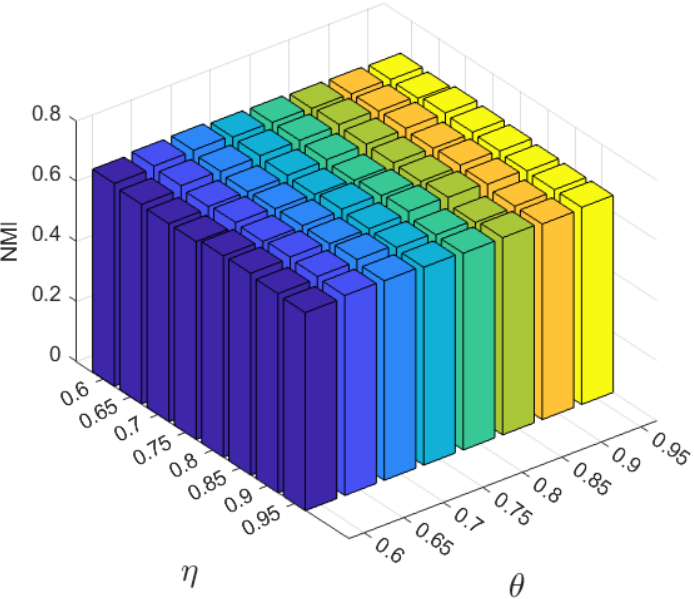}}
            \hspace{0.001\linewidth}
    \subfigure[ISOLET]{\label{para_ISOLET}
    \includegraphics[width=0.23\linewidth]{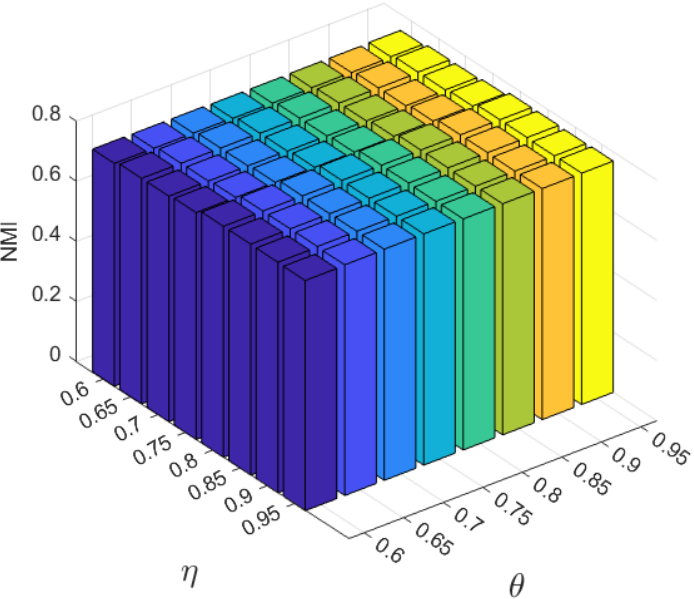}}
            \hspace{0.001\linewidth}
    \subfigure[USPS]{\label{para_USPS}
    \includegraphics[width=0.23\linewidth]{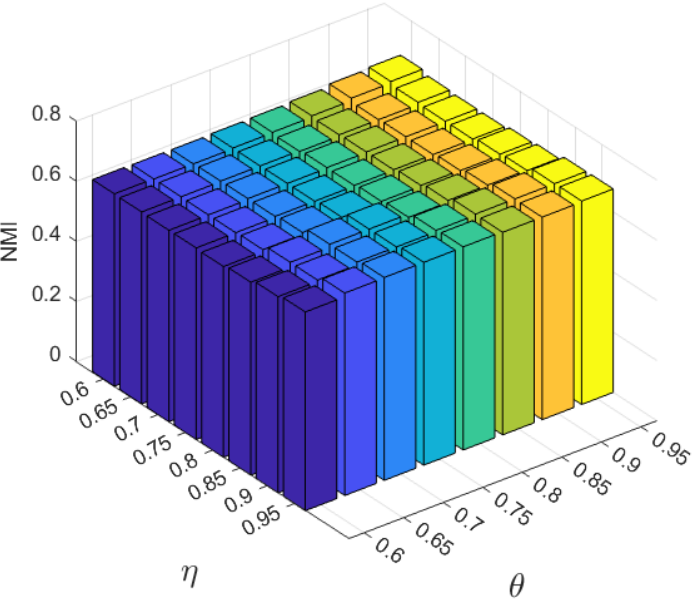}}
    
    \caption{Clustering performances with respect to NMI with varying $\eta$ and $\theta$.}
    \label{para}
\end{figure*}

\begin{figure*}
    \centering
    \subfigure[IS]{\label{NWCA_IS}
    \includegraphics[width=0.23\linewidth]{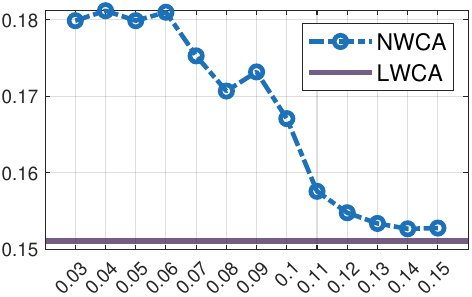}}
            \hspace{0.001\linewidth}
    \subfigure[SPF]{\label{NWCA_SPF}
    \includegraphics[width=0.23\linewidth]{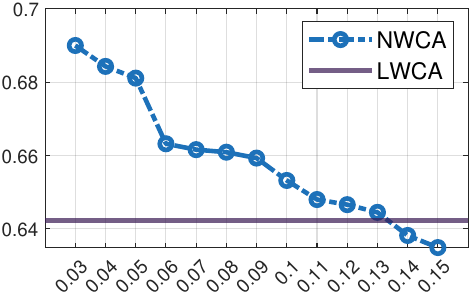}}
            \hspace{0.001\linewidth}
    \subfigure[ODR]{\label{NWCA_ODR}
    \includegraphics[width=0.23\linewidth]{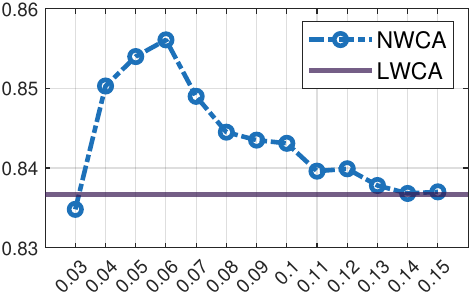}}
            \hspace{0.001\linewidth}
    \subfigure[USPS]{\label{NWCA_USPS}
    \includegraphics[width=0.23\linewidth]{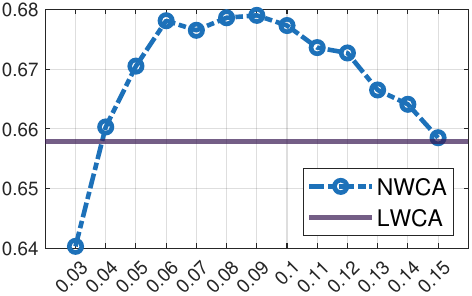}}
    \caption{Comparison of NWCA and LWCA on NMI Index and the horizontal axis represents the hyper-parameter $\lambda$.}
    \label{NWCA_vs_LWCA}
\end{figure*}

\begin{figure*}
    \includegraphics[width=1\linewidth]{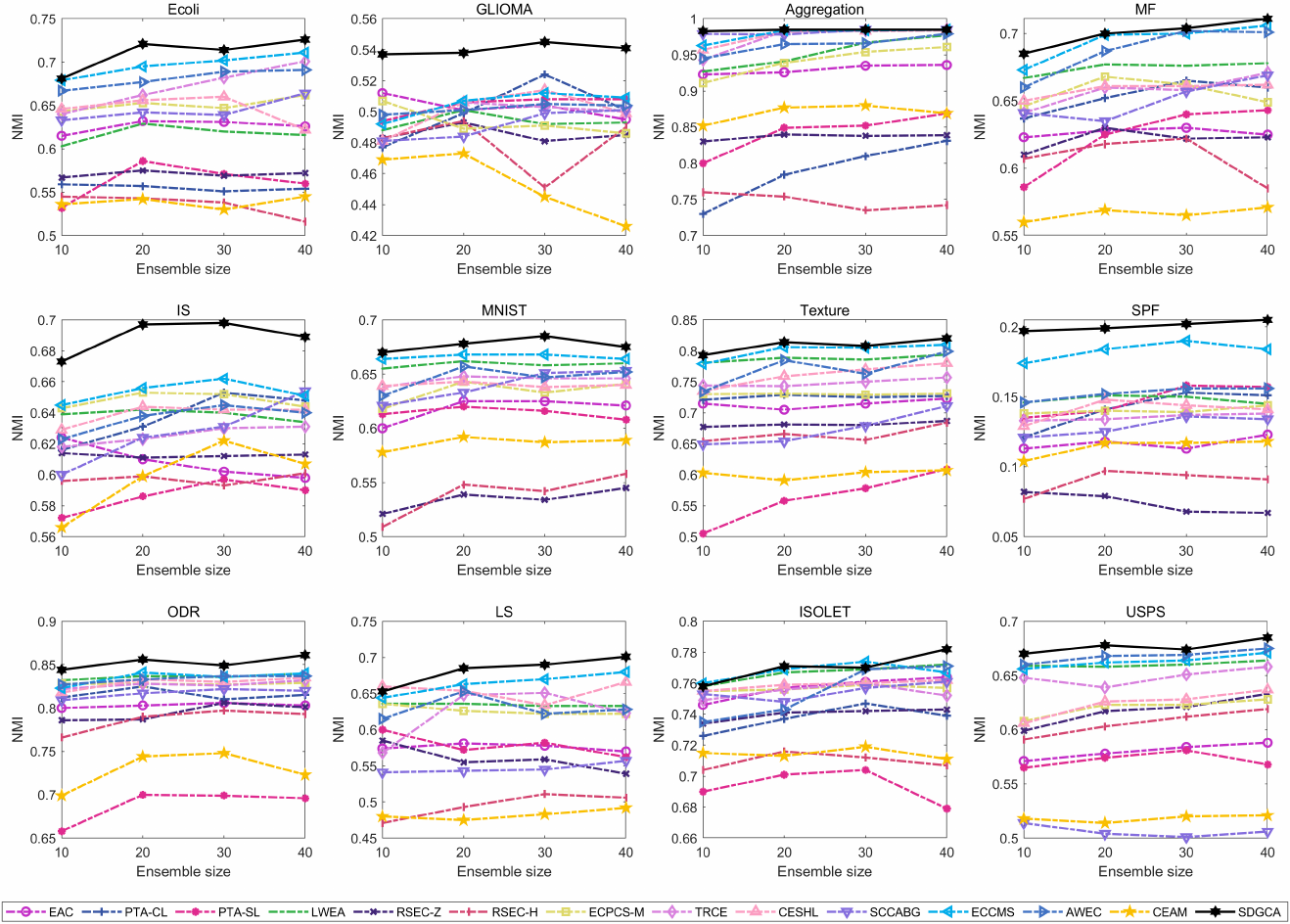}
    \caption{Clustering performances of different algorithms by varying ensemble size w.r.t. NMI.}
    \label{EnSize}
\end{figure*}

\subsection{Compared Methods}
We compared our method with the following approaches.
\begin{enumerate}
    \item \textbf{EAC} (\textit{TPAMI}, 2005) \cite{1432715}: EAC (Evidence Accumulation Clustering) combines multiple clustering results as independent evidence, generating a new similarity matrix, and then applies hierarchical clustering to obtain the final consistent partition.

    \item \textbf{PTA-CL/PTA-SL} (\textit{TKDE}, 2016) \cite{7337436}: the Probability Trajectories Aggregation (PTA) methods for robust ensemble clustering. PTA-CL, PTA-SL are two variants using complete-link and single-link agglomerative clustering respectively.

    \item \textbf{LWEA} (\textit{TCYB}, 2018) \cite{7932479}: the paper introduces Locally Weighted Ensemble Clustering method, which improves clustering by weighting the co-association matrix locally based on ensemble-driven cluster uncertainty. 

    \item \textbf{RSEC-Z/RSEC-H} (\textit{TKDD}, 2019) \cite{10.1145/3278606}: RSEC (Robust Spectral Ensemble Clustering) combines low-rank representation learning with spectral clustering to achieve robust ensemble clustering. RSEC-Z and RSEC-H obtain clustering results from the consensus partition matrix and the low-rank representation matrix respectively.

    \item \textbf{ECPCS-M} (\textit{TSMC-S}, 2021) \cite{8525437}: this method uses random walks to propagate cluster similarities in a graph. It achieves consensus clustering by grouping clusters into meta-clusters and assigning objects based on majority voting within these meta-clusters.

    \item \textbf{TRCE} (\textit{AAAI}, 2021) \cite{Zhou_Du_Shen_Li_2021}: TRCE (Tri-level Robust Clustering Ensemble) improves clustering robustness by handling base clustering, graph, and instance levels using multiple graph learning and self-paced learning for better accuracy and stability.

    \item \textbf{CESHL} (\textit{Information Fusion}, 2022) \cite{ZHOU2022171}: Hypergraphs are dynamically learned to improve clustering robustness from base clustering results, evaluate cluster quality impose constraints for clear clustering structures.

    \item \textbf{SCCABG} (\textit{TKDD}, 2023) \cite{10.1145/3564701}: SCCABG (Self-paced Adaptive Bipartite Graph Learning for Consensus Clustering) constructs an initial bipartite graph from base clustering results and iteratively refines it using self-paced learning to gradually include more reliable data.

    \item \textbf{ECCMS} (\textit{TNNLS}, 2023) \cite{10061157}: this methood improves clustering by enhancing the co-association matrix. It extracts high-confidence information from base clusterings to refine the CA matrix.

    \item \textbf{AWEC-H} (\textit{AAAI}, 2024) \cite{Xu_Li_Duan_2024}: AWEC (Adaptive Weighted Ensemble Clustering) enhances ensemble clustering by integrating high-order topological information to improve the CA matrix. It uses adaptive weights to combine multi-order connection matrices, learning an optimal connection matrix. Hierarchical clustering algorithm is applied on the CA matrix to generate the final result.

    \item \textbf{CEAM} (\textit{TKDE}, 2024) \cite{10238807}: this method refines base clustering results iteratively by constructing a multiplex graph. It uses manifold ranking for diffusion, which enhances the quality of base results through iterative updates. The final consensus clustering result is obtained from the refined multiplex.
\end{enumerate}

\subsection{Clustering Performance Comparison}
Table \ref{NMI}, \ref{ARI}, \ref{F_score} report the performance of the proposed method against other methods, the bolded values represent the optimal performance achieved on the respective datasets, while the underlined values indicate the second-best performance.

We observe that almost all methods surpass the average results produced by $k$-means, thereby validating the effectiveness of assembling the base clustering together. Moreover, it is evident that the proposed SDGCA method consistently outperforms all compared SOTA methods, showing significant improvements over the second-best method. For instance, on the SPF (D8) dataset, SDGCA improves the NMI, ARI, and F-score metrics by 8\%, 7\%, and 11\%, respectively. As for LS (D10) in metric ARI, the average $k$-means can only achieve 0.235, while ECCMS improves it to 0.616. Nonetheless, our method can further enhance it to 0.658, achieving the best performance among all methods.

In addition, not all compared methods exhibit stable performance. For instance, PTA-SL shows a significant performance drop on the MNIST (D6) and LS (D10) datasets compared to other methods. In contrast, our method exhibits robustness by effectively leveraging both similarity and dissimilarity information. When there is an error in similarity (or dissimilarity) information, the dissimilarity (or similarity) information compensates for it, providing a balancing effect.

It is noteworthy to compare the proposed method with the classical EAC method. The EAC method simply averages the co-occurrence relationships of sample pairs, thereby losing important information about cluster significance and the manifold structure of the data. However, the CA matrix provides a means to integrate multiple sources of clustering information, leading to a substantial improvement over ordinary $k$-means. Our method can compensate for the lost information, allowing for a more detailed characterization of the CA matrix, and thus achieving more reliable results.

Lastly, our method consistently surpasses the best $k$-means results on most datasets, and in cases where it does not, it remains very close to the optimal performance. This addresses the instability associated with random $k$-means initialization, providing decision-makers with highly reliable and robust outcomes.
\begin{figure*}[h]
    \centering
    \subfigure[CA matrix]{\label{Agg_CA}
    \includegraphics[width=0.182\linewidth]{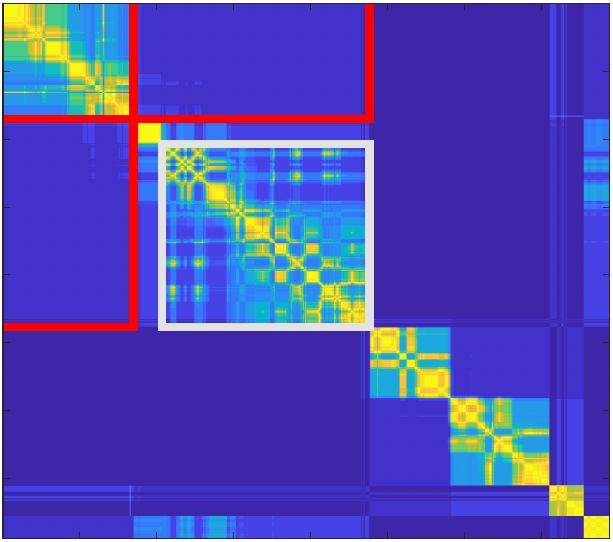}}
    \subfigure[NWCA matrix]{\label{Agg_NWCA}
    \includegraphics[width=0.182\linewidth]{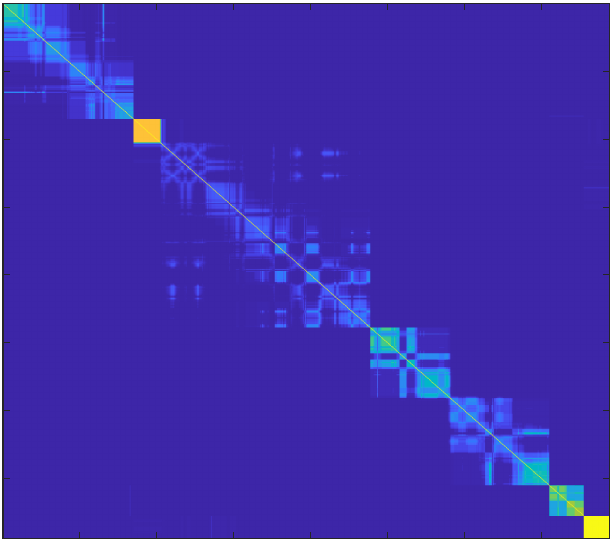}}
    \subfigure[Similarity matrix $\mathbf{S}^*$]{\label{Agg_S}
    \includegraphics[width=0.182\linewidth]{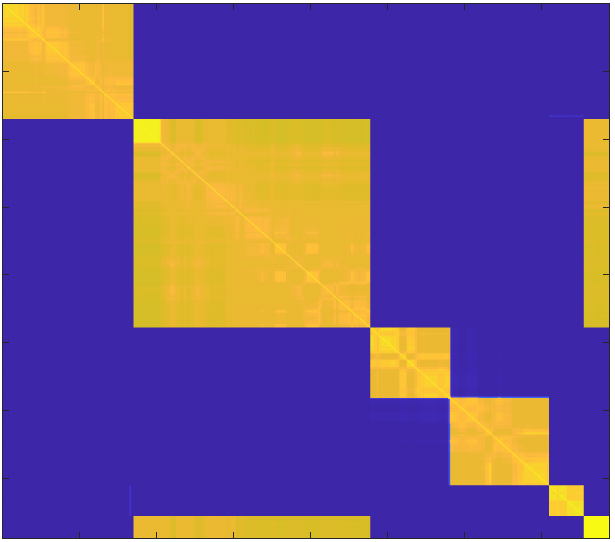}}
    \subfigure[Dissimilarity matrix $\mathbf{D}^*$]{\label{Agg_D}
    \includegraphics[width=0.182\linewidth]{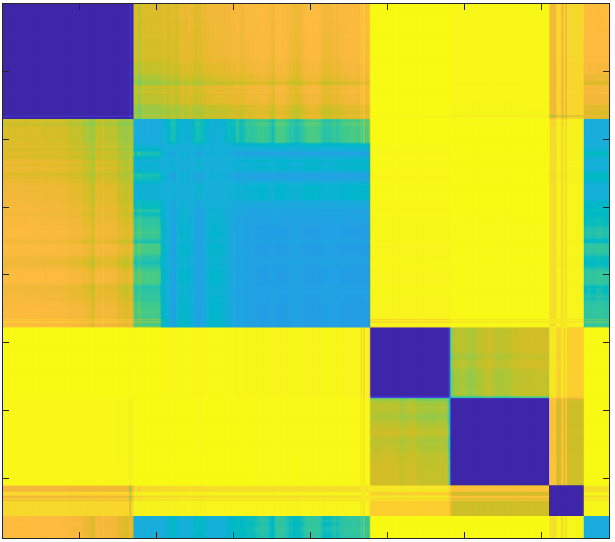}}
    \subfigure[Refined adjacency matrix $\mathbf{W}^*$]{\label{Agg_W}
    \includegraphics[width=0.206\linewidth]{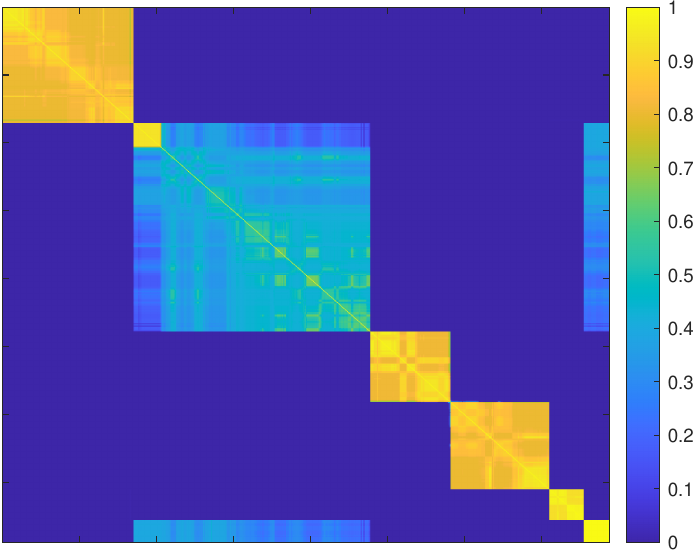}}
    
    \caption{Visualization of Aggregation dataset on different intermediate matrices and final adjacency matrix.}
    \label{vis}
\end{figure*}

\begin{figure*}
    \includegraphics[width=1\linewidth]{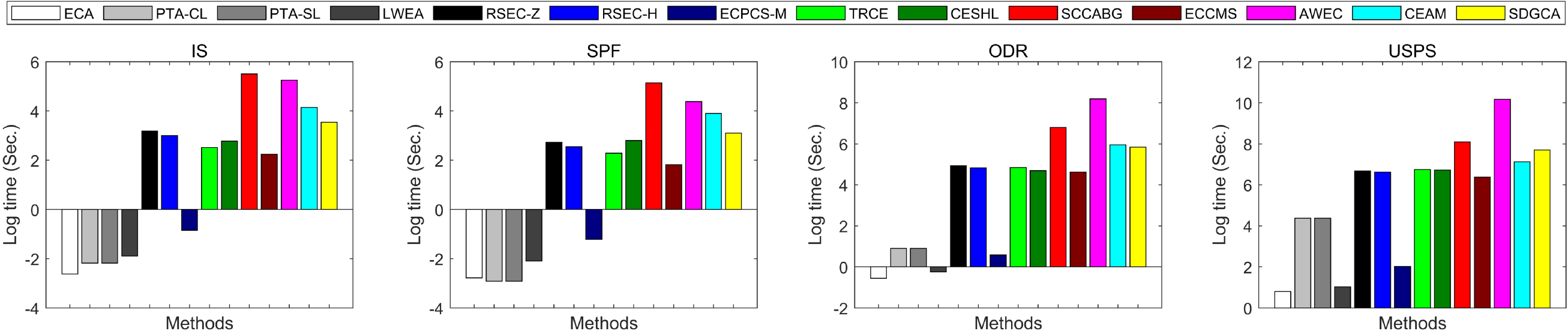}
    \caption{The runtime of different methods on various datasets.}
    \label{time}
\end{figure*}

\subsection{Ablation Study}

To further validate the effectiveness of our model, we conduct ablation experiments in this section. First, we directly use $\mathbf{S}^*$  as the adjacency matrix and perform hierarchical clustering to obtain the results, which we denote as “Only $\mathbf{S}^*$”. Next, we observe the performance decline when using only the NWCA matrix as input. Finally, we need to verify whether our manifold learning is essential. We assess the impact of removing $\mathrm{tr}\left( \mathbf{S}^{*\mathrm{T}}\mathbf{LS}^* \right)$ and $\mathrm{tr}\left( \mathbf{D}^{*\mathrm{T}}\mathbf{LD}^* \right)$ from Eq.(\ref{proposed_method}) individually, as well as removing both simultaneously, denoted as “w/o $\mathrm{tr}\left( \mathbf{D}^{*\mathrm{T}}\mathbf{LD}^* \right)$”, “w/o $\mathrm{tr}\left( \mathbf{D}^{*\mathrm{T}}\mathbf{LD}^* \right)$” and “w/o $\mathrm{tr}\left( \mathbf{S}^{*\mathrm{T}}\mathbf{LS}^* \right) + \mathrm{tr}\left( \mathbf{D}^{*\mathrm{T}}\mathbf{LD}^* \right)$” respectively.

From Table \ref{Ablation_NMI}, \ref{Ablation_ARI}, \ref{Ablation_F}, it can be observed that except for the Ecoli (D1), Texture (D7) and USPS (D12) datasets, the removal of different modules leads to varying degrees of performance degradation in other datasets. Although using “only $\mathbf{S}^*$” on the Ecoli (D1) dataset can surpass the proposed method, it is evident from other datasets that $\mathbf{S^*}$ is highly unstable and the results deviate significantly from the ground truth. Notably, for the Texture (D7) and USPS (D12) dataset, the ablation of certain modules does not result in performance improvement or decline. This is because the elements learned in $\mathbf{S}^*$ are minimal, leading to very limited adjustments to the adjacency matrix. Therefore, this experiments demonstrate that the proposed method is effective across all modules.

\subsection{Parameter Study}
Our model primarily involves three parameters: $\lambda$ in NWCA, the threshold $\theta$ for generating the Laplacian matrix from the high-confidence matrix, and the threshold $\eta$ for obtaining the similarity matrix. The analysis of the parameter $\lambda$ is shown in Fig. \ref{NWCA_vs_LWCA} (the parameter analysis of all 12 datasets are shown in Appendix). It can be seen that when $\lambda$ is set between 0.06 and 0.09, relatively good results can be achieved. Even with a fixed $\lambda$, NWCA outperforms LWCA on most datasets. Fig. \ref{para} presents the performance analysis using grid search for different values of $\theta$ and $\eta$, both varying within the range 
$\{0.6,0.65,0.7,\cdots,0.95\}$. It can be observed that our model exhibits only slight variations on the IS (D5), and SPF (D8) datasets for different parameter values, while maintaining consistent performance on other datasets. This indicates that our model is not sensitive to hyper-parameters and demonstrates strong robustness.

\subsection{Influence of Ensemble Size}
Fig. \ref{EnSize} reports the performance of all methods with different numbers of base clusterings. It can be observed that, on most datasets, the performance of our method shows an increasing trend with more base clusterings. Except for the LS and ISOLET datasets, our method consistently outperforms the compared methods across different ensemble sizes, demonstrating robustness. Notably, most methods achieve their optimal performance at an ensemble size of 40, but still do not perform as well as our method at an ensemble size of 10, as seen in the GLIOMA, IS, and ODR datasets.

\subsection{Visualization}
Fig. \ref{vis} illustrates the similarity matrix, dissimilarity matrix, and the final adjacency matrix learned by our model, compared with the original CA matrix. It is evident that NWCA weakens the noise, i.e., erroneous edges, highlighted by the red rectangles in the CA matrix. However, this makes the edges connecting sample points sparse, as indicated by the grey rectangles in the figure. To address this, we separately learn the $\mathbf{S}^*$ and $\mathbf{D}^*$ matrices. On one hand, we propagate the confidence of high-reliability sample pairs to achieve a good manifold structure. On the other hand, we use random walks to measure the confidence of samples being completely unconnected. Finally, we obtain a more reliable adjacency matrix (i.e., $\mathbf{W}^*$).

\subsection{Running Time Comparison}
Finally, we tested the runtime of the proposed method, as shown in Fig. \ref{time} (the runtime of all 12 datasets are shown in Appendix \ref{AppC}). It is observed that since the EAC, LWEA, PTA, and ECPCS methods do not involve iterative optimization, there is a gap between our method and these methods, but comparatively, our method significantly enhances clustering performance, which is worthwhile in non-real-time scenarios. In contrast to more recent methods, our approach generally delivers satisfactory results. Compared to SCCABG and AWEC, our method achieves better performance while saving a considerable amount of time. 
Overall, the performance improvement achieved by our method at the expense of time is acceptable.

\section{Conclusion}\label{Conclusion}
In this paper, we propose a novel approach to ensemble clustering through the development of the Similarity and Dissimilarity Guided Co-association matrix. We first introduce normalized ensemble entropy to explore the intrinsic qualities of clusters, establishing more precise similarity relationships. Then, we utilize random walks to build high-order connectivity between clusters and assess the credibility of direct dissimilarity among samples. Finally, adversarial learning is employed to adjust the original adjacency matrix. Extensive experiments demonstrate that our method surpasses existing state-of-the-art ensemble clustering methods significantly, showcasing its superiority.



\bibliographystyle{IEEEtran}
\bibliography{references}

\newpage\onecolumn
\appendices
\section{Convergence of the proposed method}\label{Convergence}
\textbf{Theorem 1}. Assume $\left\{ \left( \mathbf{Z}_k,\Gamma _k, \Lambda _k \right) \right\} $ is a sequence in $k$-th iteration calculated by ADMM. If $\left\{ \left(  \Gamma_k, \Lambda_k \right) \right\}, \mathrm{tr}(\mathbf{S}^{*\mathrm{T}}\mathbf{D}^*)$ are bounded and 
\begin{equation}\label{assum}
    \sum_{k=0}^\infty\left(\|\Lambda_{k+1}-\Lambda_k\|_\mathrm{F}^2+\|\Gamma_{k+1}-\Gamma_k\|_\mathrm{F}^2\right)<\infty.
\end{equation}
Then sequence $\left\{ \left( \mathbf{Z}_k,\Lambda _k, \Gamma _k \right) \right\} $ will converge and any accumulation point of $\mathbf{Z}_k$ satisfies the KKT condition.

\textit{Proof}. First, it is observed that
\begin{equation}
    \begin{aligned}
        \mathcal{L}(\mathbf{Z}, \Lambda, \Gamma) &=\mathrm{tr}\left( \mathbf{S}^{*\mathrm{T}}\mathbf{D}^* \right)+  \mathrm{tr}\left( \mathbf{S}^{*\mathrm{T}}\mathbf{LS}^* \right)+   \mathrm{tr}\left( \mathbf{D}^{*\mathrm{T}}\mathbf{LD}^* \right) 
        \\
        &\ \ \ +\mathrm{tr}\left( \Lambda^{\mathrm{T}}\left( \mathbf{S}^*-\mathbf{E} \right) \right) +\frac{\gamma_1}{2}\left\| \mathbf{S}^*-\mathbf{E} \right\| _{\mathrm{F}}^{2}
        \\
        &\ \ \ +\mathrm{tr}\left( \Gamma^{\mathrm{T}}\left(\mathbf{D}^*- \mathbf{F} \right) \right) +\frac{\gamma_2}{2}\left\| \mathbf{D}^*- \mathbf{F} \right\| _{\mathrm{F}}^{2}
        \\
        &=\mathrm{tr}\left( \mathbf{S}^{*\mathrm{T}}\mathbf{D}^* \right)+  \mathrm{tr}\left( \mathbf{S}^{*\mathrm{T}}\mathbf{LS}^* \right)+   \mathrm{tr}\left( \mathbf{D}^{*\mathrm{T}}\mathbf{LD}^* \right)
        \\ 
        &\ \ \ +\frac{\gamma _1}{2}\lVert \mathbf{S}^*-\mathbf{E}+\frac{\Lambda}{\gamma _1} \rVert _{\mathrm{F}}^{2}-\frac{1}{2\gamma _1}\lVert \Lambda \rVert _{\mathrm{F}}^{2}
        \\
        &\ \ \ +\frac{\gamma _2}{2}\lVert \mathbf{D}^*-\mathbf{F}+\frac{\Gamma}{\gamma _2} \rVert _{\mathrm{F}}^{2}-\frac{1}{2\gamma _2}\lVert \Gamma \rVert _{\mathrm{F}}^{2}
    \end{aligned}
\end{equation}
is bounded below, this follows that $\left\{ \left(  \Gamma_k, \Lambda_k \right) \right\}, \mathrm{tr}(\mathbf{S}^{*\mathrm{T}}\mathbf{D}^*)$ are bounded and Laplacian matrix $\mathbf{L}$ is positive semi-definite. Besides, $\mathcal{L}(\cdot)$ is strongly convex w.r.t each variable of $\mathbf{S}^*,\mathbf{D}^*,\mathbf{E},\mathbf{F}$, which can be seen in \textbf{Lemma 1}. Consequently, we have 
\begin{equation}\label{strcvx}
    \mathcal{L}\left( \mathbf{S}^*+\Delta \mathbf{S}^* \right) -\mathcal{L}\left( \mathbf{S}^* \right) \ge \partial _{\mathbf{S}^*}\mathcal{L}\left( \mathbf{S}^* \right) ^{\mathrm{T}}\Delta \mathbf{S}^*+\gamma _1\lVert \Delta \mathbf{S}^* \rVert _{\mathrm{F}}^{2}.
\end{equation}

For $(\mathbf{S}^*)^*$ to be the minimizer of $\mathcal{L}(\mathbf{S}^*)$, it follows 
\begin{equation}\label{mini}
    \partial _{\mathbf{S}^*}\mathcal{L}\left( \left( \mathbf{S}^* \right) ^* \right) ^{\mathrm{T}}\Delta \mathbf{S}^*\ge 0.
\end{equation}

Combining Eq. (\ref{strcvx}) and Eq. (\ref{mini}) and $\mathbf{S}^*_{k+1}$ is a minimizer of $\mathcal{L}(\mathbf{S}^*)$ at the $k$-iteration,
\begin{equation}\label{LS}
    \mathcal{L}\left( \mathbf{S}_{k}^{*} \right) -\mathcal{L}\left( \mathbf{S}_{k+1}^{*} \right) \ge \gamma _1\lVert \mathbf{S}_{k}^{*}-\mathbf{S}_{k+1}^{*} \rVert _{\mathrm{F}}^{2}.
\end{equation}

Similarly, we have
\begin{equation}
    \mathcal{L}\left( \mathbf{D}_{k}^{*} \right) -\mathcal{L}\left( \mathbf{D}_{k+1}^{*} \right) \ge \gamma _2\lVert \mathbf{D}_{k}^{*}-\mathbf{D}_{k+1}^{*} \rVert _{\mathrm{F}}^{2}.
\end{equation}
\begin{equation}
    \mathcal{L}\left( \mathbf{E}_k \right) -\mathcal{L}\left( \mathbf{E}_{k+1} \right) \ge \gamma _1\lVert \mathbf{E}_k-\mathbf{E}_{k+1} \rVert _{\mathrm{F}}^{2}.
\end{equation}
\begin{equation}\label{LF}
    \mathcal{L}\left( \mathbf{F}_k \right) -\mathcal{L}\left( \mathbf{F}_{k+1} \right) \ge \gamma _2\lVert \mathbf{F}_k-\mathbf{F}_{k+1} \rVert _{\mathrm{F}}^{2}.
\end{equation}

Let $\nu =\min \{1, \gamma_1, \gamma_2\}$ and combine Eqs. (\ref{LS}-\ref{LF}), we get
\begin{equation}
    \begin{aligned}
        \mathcal{L}\left( \mathbf{Z}_k,\Lambda _k,\Gamma _k \right) -\mathcal{L}\left( \mathbf{Z}_{k+1},\Lambda _{k+1,}\Gamma _{k+1} \right) =&\mathcal{L}( \mathbf{Z}_k,\Lambda _k,\Gamma _k ) -\mathcal{L}\left( \mathbf{Z}_{k+1},\Lambda _k,\Gamma _k \right) +\mathcal{L}\left( \mathbf{Z}_{k+1},\Lambda _k,\Gamma _k \right)-\mathcal{L}\left( \mathbf{Z}_{k+1},\Lambda _{k+1},\Gamma _{k+1} \right) 
        \\
        \ge& \nu \lVert \mathbf{Z}_k-\mathbf{Z}_{k+1} \rVert _{\mathrm{F}}^{2}-\frac{1}{\rho \gamma _1}\lVert \Lambda _k-\Lambda _{k+1} \rVert _{\mathrm{F}}^{2}-\frac{1}{\rho \gamma _2}\lVert \Gamma _k-\Gamma _{k+1} \rVert _{\mathrm{F}}^{2} 
        \\
        \ge& \nu \lVert \mathbf{Z}_k-\mathbf{Z}_{k+1} \rVert _{\mathrm{F}}^{2}-\frac{1}{\nu \rho}\left( \lVert \Lambda _k-\Lambda _{k+1} \rVert _{\mathrm{F}}^{2}+\lVert \Gamma _k-\Gamma _{k+1} \rVert _{\mathrm{F}}^{2} \right) .
    \end{aligned}
\end{equation}

Recalling that $\mathcal{L}(\mathbf{Z},\Lambda,\Gamma)$ is bounded below, we have
\begin{equation}
        \sum_{k=0}^{\infty}{\nu \lVert \mathbf{Z}_k-\mathbf{Z}_{k+1} \rVert _{\mathrm{F}}^{2}}-\sum_{k=0}^{\infty}{\frac{1}{\nu \rho}\left( \lVert \Lambda _k-\Lambda _{k+1} \rVert _{\mathrm{F}}^{2}+\lVert \Gamma _k-\Gamma _{k+1} \rVert _{\mathrm{F}}^{2} \right)} < \infty.
\end{equation}

In the previous we have assumed the second term is bounded (in Eq. (\ref{assum})), so we can immediately get
\begin{equation}
    \sum_{k=0}^{\infty}{\nu \lVert \mathbf{Z}_k-\mathbf{Z}_{k+1} \rVert _{\text{F}}^{2}}<\infty,
\end{equation}
which leads to the convergence of $\mathbf{Z}_k$, i.e., $\mathbf{Z}_{k+1}-\mathbf{Z}_k\rightarrow 0$. Similarly, $\left( \Lambda _{k+1},\Gamma _{k+1} \right) -\left( \Lambda _k,\Gamma _k \right) \rightarrow 0 $ can be directly derived from Eq. (\ref{assum}).

Next, we proceed to prove that the accumulation points of $\mathbf{Z}_k$ satisfy the KKT condition. The KKT condition of problem (\ref{intermediate}) is given as
\begin{equation}
    \mathbf{D}^*+\Lambda +2\mathbf{LS}^*=0,
\end{equation}
\begin{equation}
    \mathbf{S}^*+\Gamma +2\mathbf{LD}^*=0,
\end{equation}
\begin{equation}\label{KKT3}
    \mathbf{S}^*-\mathbf{E}=0,
\end{equation}
\begin{equation}\label{KKT4}
    \mathbf{D}^*-\mathbf{F}=0,
\end{equation}
\begin{equation}\label{KKT5}
    0\le \mathbf{E}\le 1, \mathbf{E}=\mathbf{E}^{\text{T}}, \mathcal{P}_{\Omega^\mathbf{S}}\left( \mathbf{E} \right) =\mathbf{S},
\end{equation}
\begin{equation}\label{KKT6}
    0\le \mathbf{F}\le 1, \mathbf{F}=\mathbf{F}^{\text{T}}, \mathcal{P}_{\Omega^\mathbf{D}}\left( \mathbf{F} \right) =\mathbf{D},
\end{equation}
where $\odot$ represents the Hadamard product, which is an element-wise multiplication operation.

To prove it, we rewrite Eqs. (\ref{S_solution}), (\ref{F1_solution}), (\ref{D_solution}), (\ref{F2_solution}) as 
\begin{equation} \label{rewrite_S}
    \left( 2\mathbf{L}+\gamma _1\mathbf{I} \right) \left( \mathbf{S}_{k+1}^{*}-\mathbf{S}_{k}^{*} \right) =\gamma _1\left( \mathbf{E}_k-\mathbf{S}_{k}^{*} \right) -\left( \mathbf{D}_{k}^{*}+\Lambda _k+2\mathbf{LS}_{k}^{*} \right) 
\end{equation}

\begin{equation} \label{rewrite_D}
    \left( 2\mathbf{L}+\gamma _2\mathbf{I} \right) \left( \mathbf{D}_{k+1}^{*}-\mathbf{D}_{k}^{*} \right) =\gamma _2\left( \mathbf{F}_k-\mathbf{D}_{k}^{*} \right) -\left( \mathbf{S}_k+\Gamma _k+2\mathbf{LD}_{k}^{*} \right) 
\end{equation}

\begin{equation} \label{rewrite_E}
    \mathbf{E}_{k+1}-\mathbf{E}_k=\mathcal{P}_{\Omega^\mathbf{S}}\left( \max \left( \min \left( \frac{\left( \mathbf{P}_k+\mathbf{P}_{k}^{\text{T}} \right)}{2},1 \right) ,0 \right) \right) -\mathbf{E}_k, \mathbf{P}_k=\mathbf{S}_{k}^{*}+\frac{\Lambda _k}{\gamma _1}
\end{equation}

\begin{equation} \label{rewrite_F}
    \mathbf{F}_{k+1}-\mathbf{F}_k=\mathcal{P}_{\Omega^\mathbf{D}}\left( \max \left( \min \left( \frac{\left( \mathbf{Q}_k+\mathbf{Q}_{k}^{\text{T}} \right)}{2},1 \right) ,0 \right) \right) -\mathbf{F}_k, \mathbf{Q}_k=\mathbf{D}_{k}^{*}+\frac{\Gamma _k}{\gamma _2}
\end{equation}

\begin{equation} \label{Lambda}
    \Lambda _{k+1}-\Lambda _k=\gamma _1\left( \mathbf{S}_{k}^{*}-\mathbf{E}_k \right) 
\end{equation}

\begin{equation} \label{Gamma}
    \Gamma _{k+1}-\Gamma _k=\gamma _2\left( \mathbf{D}_{k}^{*}-\mathbf{F}_k \right) 
\end{equation}

Since we have already proven the convergence of $\left\{ \left( \mathbf{Z}_k,\Lambda _k, \Gamma _k \right) \right\} $, it is evident that $\mathbf{S}_{k}^{*}-\mathbf{E}_k \rightarrow 0,\mathbf{D}_{k}^{*}-\mathbf{F}_k\rightarrow 0$ in Eqs. (\ref{Lambda}) and (\ref{Gamma}), which supplys the sufficient condition for Eqs. (\ref{KKT3}) and (\ref{KKT4}). Combining it with Eqs (\ref{rewrite_S}), (\ref{rewrite_D}), we obtain the first and second equalities in the KKT condition. For the constraints of $\mathbf{E}$ and $\mathbf{F}$ in Eqs. (\ref{KKT5}), (\ref{KKT6}), these can be easily derived from Eqs (\ref{rewrite_E}) and (\ref{rewrite_F}). This completes the proof.

\textbf{Lemma 1}. The lagrangian function $\mathcal{L}(\cdot)$ is strongly convex w.r.t each variable of $\mathbf{S}^*,\mathbf{D}^*,\mathbf{E},\mathbf{F}$. 

\textit{Proof}. Since $\mathbf{S}^*$ and $\mathbf{D}^*$, as well as $\mathbf{E}$ and $\mathbf{F}$, share the same form, we only need to prove one of them. 

Obviously, $\mathcal{L}(\mathbf{E})=\frac{\gamma _1}{2}\lVert \mathbf{E}-\mathbf{M}_0 \rVert +c_0$ ($\mathbf{M}_0$ and $c_0$ denoted as a constant matrix and a constant) is a strongly convex function. For $\mathcal{L}(\mathbf{S}^*)$, it can be written as $\mathcal{L}(\mathbf{S}^*)=\mathrm{tr}\left( \mathbf{S}^{*\mathrm{T}}\mathbf{M}_1 \right) +\mathrm{tr}\left( \mathbf{S}^{*\mathrm{T}}\mathbf{LS}^* \right) +\frac{\gamma _1}{2}\lVert \mathbf{S}^*-\mathbf{M}_2 \rVert _{\mathrm{F}}^{2}+c_1$. Since $\lVert \mathbf{S}^*-\mathbf{M}_2 \rVert _{\mathrm{F}}^{2}$ is a positive definite quadratic function of $\mathbf{S}^*$, and $\mathrm{tr}\left( \mathbf{S}^{*\mathrm{T}}\mathbf{M}_1 \right) +\mathrm{tr}\left( \mathbf{S}^{*\mathrm{T}}\mathbf{LS}^* \right)$ is a convex function, according to the definition of a strongly convex function, $\mathcal{L}(\mathbf{S}^*)$ is also strongly convex.

\newpage
\section{Analysis of $\lambda$ parameter across all datasets}\label{AppB}
\begin{figure*}[htbp]
    \centering
    \subfigure[Ecoli]{
    \includegraphics[width=0.23\linewidth]{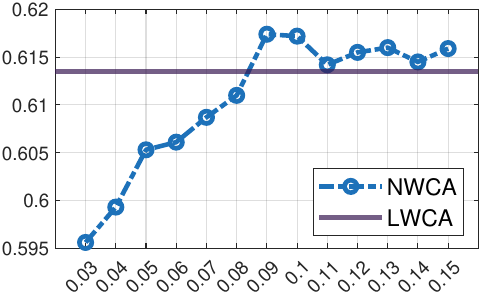}}
    \hspace{0.001\linewidth}
    \subfigure[GLIOMA]{
    \includegraphics[width=0.23\linewidth]{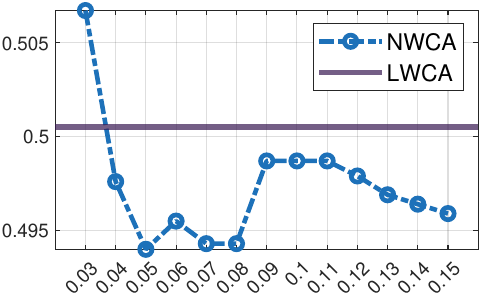}}
    \hspace{0.001\linewidth}
    \subfigure[Aggregation]{
    \includegraphics[width=0.23\linewidth]{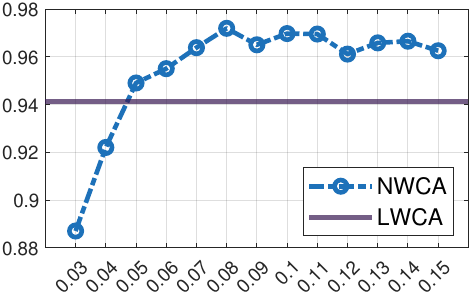}}
    \hspace{0.001\linewidth}
    \subfigure[MF]{
    \includegraphics[width=0.23\linewidth]{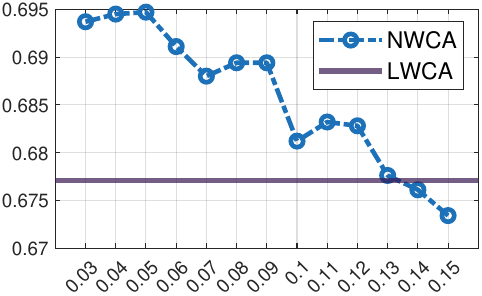}}
        \hspace{0.001\linewidth}
    \subfigure[IS]{
    \includegraphics[width=0.23\linewidth]{fig/NWCA/NWCA_IS.pdf}}
        \hspace{0.001\linewidth}
    \subfigure[MNIST]{
    \includegraphics[width=0.23\linewidth]{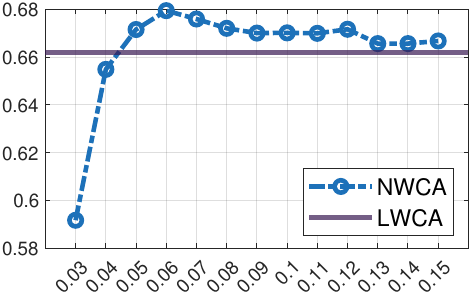}}
        \hspace{0.001\linewidth}
    \subfigure[Texture]{
    \includegraphics[width=0.23\linewidth]{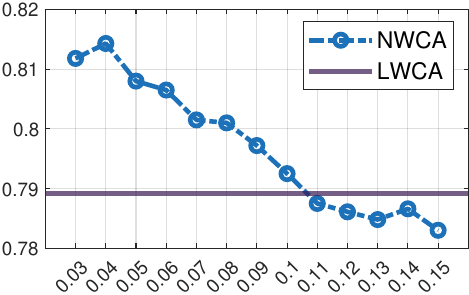}}
        \hspace{0.001\linewidth}
    \subfigure[SPF]{
    \includegraphics[width=0.23\linewidth]{fig/NWCA/NWCA_SPF.pdf}}
            \hspace{0.001\linewidth}
    \subfigure[ODR]{
    \includegraphics[width=0.23\linewidth]{fig/NWCA/NWCA_ODR.pdf}}
            \hspace{0.001\linewidth}
    \subfigure[LS]{
    \includegraphics[width=0.23\linewidth]{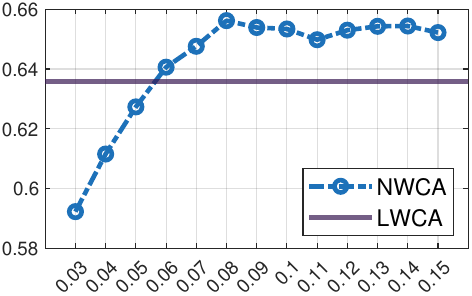}}
            \hspace{0.001\linewidth}
    \subfigure[ISOLET]{
    \includegraphics[width=0.23\linewidth]{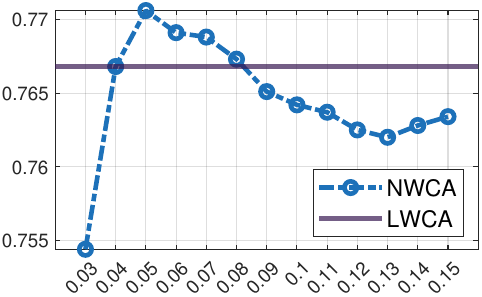}}
            \hspace{0.001\linewidth}
    \subfigure[USPS]{
    \includegraphics[width=0.23\linewidth]{fig/NWCA/NWCA_USPS.pdf}}
    
    \caption{Comparison of NWCA and LWCA on the NMI Index.}
\end{figure*}

\newpage
\section{The runtime of different methods across all datasets}\label{AppC}
\begin{figure*}[htbp]
    \includegraphics[width=1\linewidth]{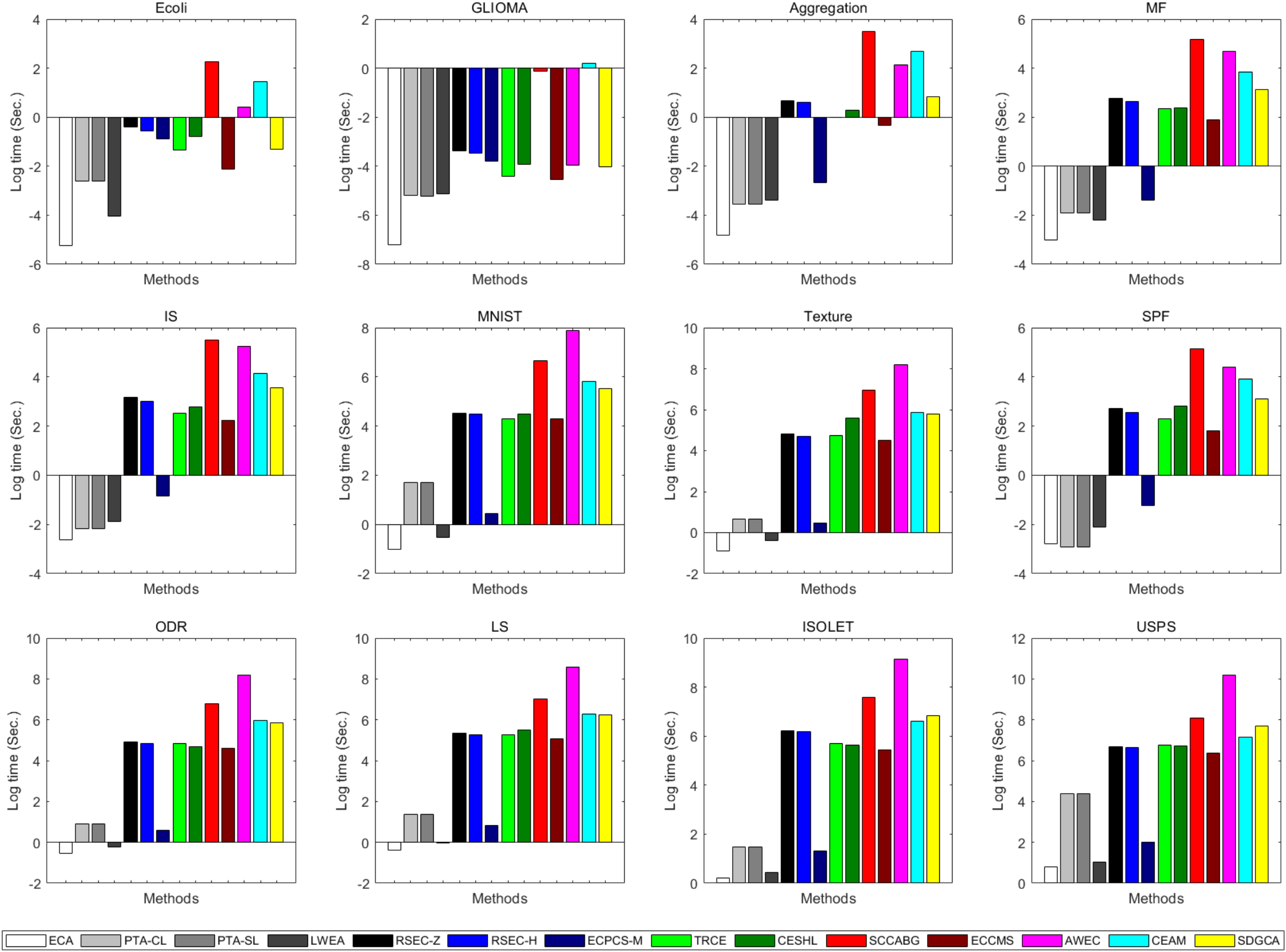}
    \caption{The runtime of different methods on various datasets.}
\end{figure*}
\twocolumn

\end{document}